\PassOptionsToPackage{main=english,russian}{babel}
\PassOptionsToPackage{T2A}{fontenc}
\documentclass[onecolumn]{komodo_test}

\usepackage{lipsum}
\usepackage{pdfpages}
\usepackage{colortbl}
\usepackage{tabulary}
\usepackage{longtable}
\usepackage{array}
\usepackage{placeins}
\usepackage{tablefootnote}
\usepackage{tcolorbox}
\usepackage{wrapfig}
\usepackage{listings}

\usepackage{breqn} 

\usepackage{pifont}
\definecolor{deeppurple}{HTML}{9e02f7}
\definecolor{forestgreen}{HTML}{2e7d43}
\usepackage{multirow}
\usepackage{tabularx}

\usepackage[utf8]{inputenc}

\usepackage{arydshln}

\usepackage{xspace} 
\usepackage{makecell} 
\usepackage{multirow}

\usepackage[framemethod=TikZ]{mdframed}
\usepackage{tcolorbox}
\usepackage{multicol}
\usepackage{dblfloatfix}

\newtcolorbox{mybox}[2][]{
  colback=white, 
  colframe=lightblue,
  fonttitle=\bfseries,
  coltitle=black,  
  title=#2, 
  #1
}

\definecolor{ayad}{RGB}{148, 156, 229} 
\definecolor{ayadsymbol}{RGB}{76, 110, 230} 
\definecolor{lightblue}{RGB}{211, 227, 252} 
\definecolor{bgblue}{RGB}{247, 250, 255} 
\newcommand*\colourcheck[1]{%
  \expandafter\newcommand\csname #1check\endcsname{\textcolor{#1}{\ding{52}}}%
}
\newcommand*\colourcross[1]{%
  \expandafter\newcommand\csname #1cross\endcsname{\textcolor{#1}{\ding{55}}}%
}
\DeclareSymbolFont{extraup}{U}{zavm}{m}{n}
\DeclareMathSymbol{\vardiamond}{\mathalpha}{extraup}{87}
\definecolor{ayadsymbol}{RGB}{76, 110, 230} 

\colourcheck{blue}
\colourcheck{green}
\colourcheck{red}

\colourcross{blue}
\colourcross{green}
\colourcross{red}
\usepackage{nicematrix}
\usepackage{comment}
\usepackage{amssymb}

\setcitestyle{number,square}

\title{\includegraphics[scale=0.07]{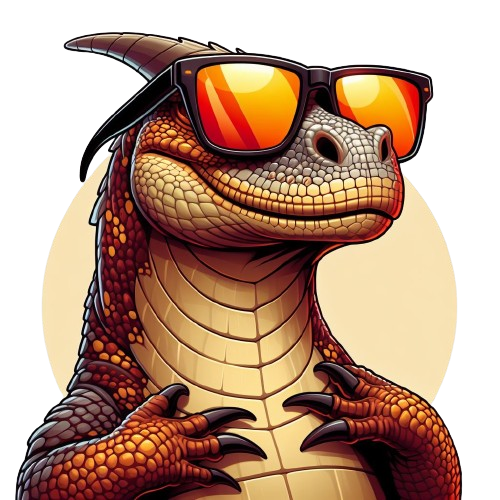}Komodo: A Linguistic Expedition into Indonesia's Regional Languages}

\multiauthors
\author{
    name={Louis Owen$^\circ$},
}

\author{
    name={Vishesh Tripathi$^\circ$}
}

\author{
    name={Abhay Kumar$^\dagger$}
}

\author{
   name={Biddwan Ahmed$^\dagger$}
}

\affiliations{
    NLP Research Team,Yellow.ai\\$^\circ$First authors;$^\dagger$Senior authors}

\date{\today}
\abstract{
The recent breakthroughs in Large Language Models (LLMs) have mostly focused on languages with easily available and sufficient resources, such as English. However, there remains a significant gap for languages that lack sufficient linguistic resources in the public domain. Our work introduces Komodo-7B, 7-billion-parameter Large Language Models designed to address this gap by seamlessly operating across Indonesian, English, and 11 regional languages in Indonesia. Komodo-7B is a family of LLMs that consist of Komodo-7B-Base and Komodo-7B-Instruct. Komodo-7B-Instruct stands out by achieving state-of-the-art performance in various tasks and languages, outperforming the benchmarks set by OpenAI’s GPT-3.5, Cohere’s Aya-101, Llama-2-Chat-13B, Mixtral-8x7B-Instruct-v0.1, Gemma-7B-it , and many more. This model not only demonstrates superior performance in both language-specific and overall assessments but also highlights its capability to excel in linguistic diversity. Our commitment to advancing language models extends beyond well-resourced languages, aiming to bridge the gap for those with limited linguistic assets. Additionally, Komodo-7B-Instruct's better cross-language understanding contributes to addressing educational disparities in Indonesia, offering direct translations from English to 11 regional languages, a significant improvement compared to existing language translation services. Komodo-7B represents a crucial step towards inclusivity and effectiveness in language models, providing  to the linguistic needs of diverse communities.
}

\renewcommand{\FONTauthors}{\bfseries\Large}

\begin{document}

\section{Introduction}

Since the introduction of transformers~\citep{NIPS2017_3f5ee243} based language model, there is a significant gap when it comes to models tailored to specific regional languages. While models like GPT-3.5~\citep{Chat-GPT} and Llama-2~\citep{touvron2023llama} excel in various tasks, their performance is primarily benchmarked in English. However, these models tend to underperform when dealing with languages other than english, on the other hand, there are multilingual models like Aya-101~\citep{aya-model}, Bactrian-X~\citep{li2023bactrianx}, Qwen-1.5~\citep{qwen15}, and Mixtral~\citep{jiang2024mixtral}, that perform well in tasks involving multiple languages. Yet, when it comes to individual languages or small regional languages with limited available data, these models lack specialized expertise. Also significant advancements have been achieved in creating datasets for pre-training and instruction tuning, such as Alpaca~\citep{alpaca}, UltraChat~\citep{ding2023enhancing}, Dolly~\citep{DatabricksBlog2023DollyV2}, OpenAssistant~\citep{köpf2023openassistant}, and LMSYS-Chat~\citep{zheng2023lmsyschat1m}. Additionally, there have been efforts to establish evaluation benchmarks like AlpacaEval~\citep{alpaca_eval} and MT-Bench. However, it's important to note that the majority of these developments have primarily focused on the English language.

Nevertheless, the way data is represented, the efficiency of tokenizers, and the overall performance in tasks related to Indonesian and regional languages lag significantly behind those of English. Even in closed-source models like GPT-3.5, GPT-4~\citep{openai2024GPT4}, and similar ones, the performance in Indonesian languages falls short when compared to their performance in English. This performance gap highlights the need for focused attention and improvement in addressing the specific challenges posed by Indonesian and regional languages in the realm of language models. So, overall, currently, there is a notable absence of high performing LLMs specifically designed for Indonesia, trained on Indonesian data and evaluated against benchmarks for Indonesia's regional languages. In response to this gap, we present Komodo-7B-Instruct, a language model with 7 billion parameters, tailored for 13 languages including Indonesian, English and 11 regional languages.

In our comparison with several big models, both those available to everyone and those with restricted access like ChatGPT, Llama-2, and Mixtral, we have achieved top-notch performance in a few tasks. We've surpassed the capabilities of various multilingual models like Cohere’s Aya-101, MBZUAI’s Bactrian-X-llama-7B, Qwen-1.5, and Mixtral-8x7B-Instruct-v0.1, across several tasks. Notably, our performance also surpasses the existing Indonesian Large Language Model (LLM) SEA-LION~\citep{sea_lion_2023} in almost every task by a significant margin. This demonstrates our model's effectiveness and superiority in handling diverse tasks and outperforming existing benchmarks.

In addition, our model shows excellent cross-language understanding, making it a valuable tool to bridge the education gap in Indonesia. With the capability to directly translate from English to 11 regional languages, or even from one regional language to others, our model surpasses the limitations of Google Translate~\citep{google-translate}, which only supports Indonesian, Javanese, and Sundanese. Importantly, our model also serves to a broader range of regional languages, ensuring accessibility for people across various regions in Indonesia.

It's worth noting that Javanese and Sundanese are predominantly spoken on the Java island, where innovation and education quality are relatively higher compared to other regions in Indonesia. Our model's support for a diverse set of regional languages ensures that educational resources and information can be more widely disseminated, contributing to a more inclusive and equitable educational landscape throughout the country.

\section{Dataset}

The dataset employed in both the pre-training and fine-tuning phases of our language model was created not only from diverse open-source datasets but also from the manually collected data for regional languages of Indonesia.  Our approach is informed by the noteworthy outcomes demonstrated by models such as Phi-1~\citep{gunasekar2023textbooks} and Phi-1.5~\citep{li2023textbooks}, indicating that achieving commendable results does not necessarily depends on vast amounts of data. Instead, a judicious selection of high-quality data has proven effective, even yielding State-of-the-Art performance under certain circumstances. As data preprocessing is a crucial step in scaling the language models, we took some preprocessing steps inspired by~\citep{rae2022scaling}, as follows:

\textbf{Repetition Removal}: Excessive repetition of words or phrases can lead to uninformative content and is a common issue in language models. To address this, we identify and remove documents with a high proportion of repeated lines or paragraphs. Utilizing multiple approaches, we calculate the fraction of duplicate content for lines and paragraphs separately, considering different n-gram sizes. This ensures a comprehensive removal of redundancy, contributing to the refinement of our dataset.

\textbf{Quality Filtering}: Implementing measures to filter out low-quality or irrelevant data, contributing to a more refined dataset. To enhance data quality, we implement straightforward heuristic filters. We exclude documents with insufficient or excessive word counts, ensuring they fall within a specific range. By applying these filters, we aim to retain high-quality, coherent text relevant to language model training.

\textbf{Deduplication}~\citep{chenghao_mou_2023_8364980}: Duplicate text across various documents can introduce redundancy in the dataset. Identifying and removing duplicate entries to prevent redundancy and ensure the uniqueness of the dataset is an important step towards maintaining high-quality data.

These preprocessing steps are integral to our commitment to data quality, aiming to enhance the effectiveness and performance of our language model.

\subsection{Pretraining \& Supervised-Fine-Tuning Data} 

Moreover, our research extends to the incorporation of Indonesian textbooks spanning grades 1 through 12, consisting a number of subjects including arts, civics, mathematics, sports, religion, local cultures, and engineering, among others. This strategic integration serves the purpose of enhancing the model's general knowledge capabilities, covering a broad array of topics including science, daily activities, and more. We've also integrated colloquial data extracted from various sources such as movie subtitles, news, informal conversations, movie reviews, poems, and more. 

Additionally, we made use of freely available datasets primarily collected in Indonesian and other regional languages, such as Javanese, Sundanese, Acehnese, and many more. We aim to ensure that our language model is well-versed not only in Indonesian but also in other regional languages. The primary objective is to improve our model’s comprehension of regional languages, as currently, no language model understands different regional languages in Indonesia. This approach helps improve the model's overall language skills and adaptability to various cultural contexts.

The intention behind this comprehensive dataset combination is to imbue the language model with a better and deep understanding of the language, encouraging adeptness in handling diverse contextual cues and promoting a more inclusive comprehension of language and its applications. 

In our dataset, we've taken inspiration from OpenHathi~\citep{openhathi} approach to include English datasets and alternate parallel data, aiming to enhance our model's understanding of code-mixed sentences. Alternate parallel, as inspired by OpenHathi, involves a unique approach to teaching cross-lingual understanding. Instead of following the traditional method of monolingual next-token prediction with translated Indonesian text, we employ a bilingual next-token prediction strategy.

This bilingual approach introduces alternate sentences in English and Indonesian. The significance lies in requiring the model to cross-lingually attend to information during next-token prediction. For instance, predicting an English token in the second sentence would necessitate attending to Indonesian tokens in the preceding sentence. We hypothesize that this approach increases alignment between English and Indonesian. Moreover, it naturally balances the exposure of the model to both languages during training, promoting a more robust understanding.

It's crucial to note that our use of alternate parallelism is not limited to English and Indonesian only; rather, it encompasses all combinations of English, Indonesian, and the 11 regional languages, including Acehnese, Balinese, Banjarese, Buginese, Dayak Ngaju, Javanese, Lampungnese, Madurese, Minangkabau, Sundanese, and Toba Batak. This inclusive approach ensures a comprehensive and diverse training set for our model, contributing to its proficiency in handling code-mixed sentences across multiple languages.

To facilitate translation between English and Indonesian whenever needed, we also leverage translation engines like Google Translate API. This additional step further enriches our dataset and supports the model's ability to comprehend and generate content in diverse linguistic scenarios.

The SFT dataset is derived from open-source data, covering a wide array of tasks. We also do manual label creation from unlabeled data , such as the creation of a Multiple-Choice Question task using the Indonesian dictionary (KBBI). The SFT dataset also incorporates responses from ChatGPT, adding nuanced and diverse perspectives to the labeled data. The meticulous curation process ensures a comprehensive and adaptable dataset, making it valuable for training models across various language-related tasks.

\begin{table}[!t]
\centering
\footnotesize
\begin{tabular}{@{}llcccc@{}}
\toprule
\textbf{Task}          & \textbf{Dataset}     & \textbf{Metric} & \textbf{Unseen Tasks}     & \textbf{Languages} \\ \midrule
\multicolumn{5}{l}{\textbf{Discriminative Tasks}}                                                             &                    \\
MCQs                   & IndoMMLU                        & Acc.            & \redcross                         & 10                  \\
Entailment             & ID-EN                           & Acc.            & \redcross                          & 2                  \\
Common Sense Reasoning & X-Copa-ID                        & Acc.            & \redcross                          & 1                  \\
Intent-Classification  & Intent-Classification            & F1-w-avg        & \bluecheck & 3                  \\
Colloquial-Detection   & Colloquial-Detection             & Acc.            & \bluecheck & 1                  \\
Sentiment-Analysis     & NusaXSenti                       & Acc.            & \bluecheck & 11                 \\
Hatespeech Detection   & ID-Hatespeech                   & Acc.            & \bluecheck & 1                  \\
\multicolumn{5}{l}{\textbf{Generative Tasks}}                                                                 &                    \\
Translation  & NusaX-MT & CHRF++ & \redcross & 13 \\
Question-Answering     & Tydiqa-ID                        & Acc.            & \redcross                          & 1                  \\
Summarisation          & IndoSum                          & Rouge-L-F1      & \redcross                          & 1                  \\ \bottomrule
\end{tabular}
\caption{Datasets considered for evaluation. Unseen Task refers to tasks entirely excluded from training, which includes the 3 discriminative tasks. The seen tasks refer to the tasks where supervised fine tuning is
performed and instances are held-out for evaluation.}
\label{tab:my-table}
\end{table}

\subsection{Benchmarking Datasets}



\textbf{IndoMMLU}~\citep{koto-etal-2023-indommlu}: The IndoMMLU dataset is designed to assess language proficiency by focusing across diverse subjects including STEM, humanities, social sciences, and more. Questions within the dataset prompt responses that cover a spectrum of topics, providing a thorough evaluation of language understanding and expression in Indonesian and regional languages.

\textbf{Indotexbook Bilingual (ID-EN) Entailment}: The Indotexbook Bilingual Entailment dataset involves paired sentences in Indonesian and English, intended for entailment analysis. It tasks the model with determining the relationship between two sentences, one in Indonesian and the other in English. The dataset facilitates evaluation for cross-lingual entailment understanding. This is manually created data (held-out set) based on the Indonesian textbooks data that we utilized during pretraining and SFT.

\textbf{X-Copa (ID)}~\citep{ponti2020xcopa}: X-Copa, specific to Indonesian (ID), presents contextualized sentence pairs to evaluate the model's common-sense reasoning abilities. The dataset includes scenarios where the model must infer the consequence or outcome of a given situation. For instance, inferring the result of a person turning on a tap. This dataset aids in assessing the model's ability to draw logical inferences.

\textbf{NusaXSenti}~\citep{winata2022nusax}: NusaXSenti focuses on sentiment analysis and classification into positive, negative, or neutral sentiments. The dataset includes sentences in Indonesian, English, and multiple regional languages, broadening the scope of sentiment analysis across diverse linguistic contexts. However, it is worth noting that during benchmarking, we’re filtering out the English subset since our goal is to measure the model’s performance on Indonesian and regional languages.

\textbf{Intent Classification}: Intent Classification involves classifying the intent behind a given sentence. The dataset provides cases of sentences where the model needs to identify the specific intent, contributing to the development and evaluation of models for intent recognition tasks. The dataset is based on BANKING77-OOS~\citep{zhang2022pretrained} data, where we sample only 5 indomain intents and 1 out-of-domain out-of-scope intent (ood-oos). Original utterances are in English, while we translate them into Indonesian, Javanese, and Sundanese.

\textbf{Colloquial Detection}: Colloquial Detection aims to classify the level of formality in a given sentence. The dataset provides tasks for the model to classify whether a given sentence is a colloquial sentence or not. We created this dataset based on the Twitter data from the emotion classification task released by IndoNLU~\citep{wilie2020indonlu}. We mark this set as the colloquial sentences. While for the formal sentences, we sampled several lines from our Indonesian textbooks pre-training data.

\textbf{ID-HateSpeech}~\citep{idhatespeech}: As the name suggests, ID-HateSpeech is designed specifically for hate speech detection. This dataset aids in evaluating the model to identify and categorize content containing hate speech, contributing to the development of robust hate speech detection systems.

\textbf{TydiQA-ID}~\citep{cahyawijaya-etal-2021-indonlg}~\citep{clark-etal-2020-tydi}: TydiQA-ID contributes to the benchmarking efforts with a focus on Indonesian language question answering. It is an extractive Question Answering dataset. This aids in evaluating the model's comprehension of historical and factual information in Indonesian.

\textbf{IndoSum}~\citep{indosum}: IndoSum is geared towards summarization tasks, providing the model with text to generate concise and informative summaries. This dataset is valuable for evaluating the model in conceptual summarization techniques.

\textbf{NusaX-MT}~\citep{winata2022nusax}: NusaX-MT involves machine translation tasks, where the model is tasked with translating sentences from one language to another. It instructs the model to translate a given sentence from one regional language to another regional language, showcasing the dataset's multilingual translation challenges. Similar to NusaXSenti, this data also consists of Indonesian, English, and 10 regional languages.

\section{Training and Experimental Setup}

\begin{figure}[t]
     \centering
     \includegraphics[width=1\textwidth]{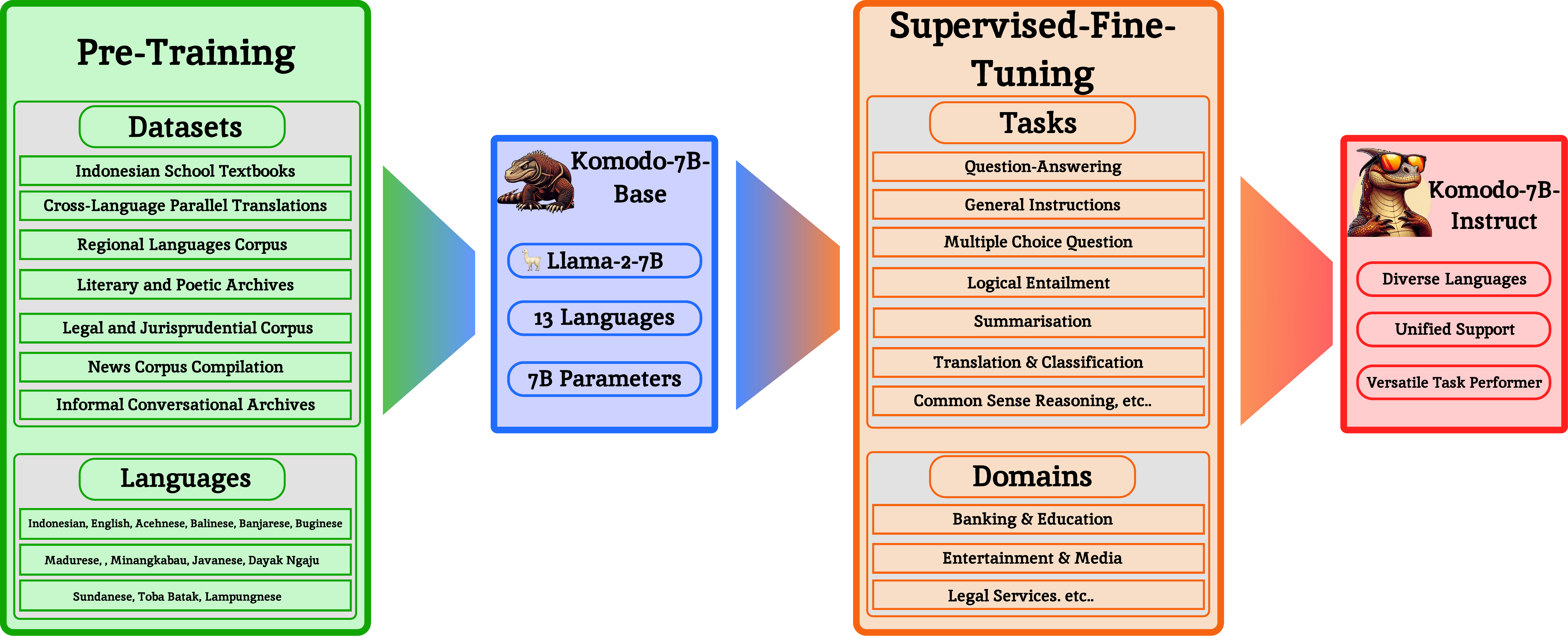}
     \caption{The Evolution of Komodo-7B-Instruct Language Model. The diagram illustrates the transformation from the Komodo-7B-Base model, initially trained on diverse datasets encompassing various languages, to the refined Komodo-7B-Instruct model through targeted Supervised Fine-Tuning (SFT) on specific tasks and domains. The journey involves strategic pretraining on comprehensive datasets, followed by fine-tuning for enhanced performance and adaptability across a spectrum of language-related challenges.}
     \label{fig:architecture-diagram}
\end{figure}

We have built our language model on top of the Llama-2 model,but with some unique adjustments for our needs.

\subsection{Expanding the Vocabulary} Recognizing the importance of linguistic diversity, we focused on enhancing our language model's proficiency in both Indonesian and regional languages. To achieve this, we systematically expanded the tokenizer's vocabulary by identifying and incorporating approximately 2,000 frequently used words specific to Indonesian and 1,000 words for regional languages that were absent in the Llama-2 model. 

The standard method for enhancing a vocabulary typically involves developing a new tokenizer and integrating it with the existing one. This technique has shown impressive results in projects like Chinese-LLaMA~\citep{Chinese-LLaMA-Alpaca} and Open-Hathi~\citep{openhathi}. The effectiveness of this strategy can be attributed to the significant linguistic distinctions between languages such as Chinese and Hindi when compared to English. In contrast, the Indonesian language employs the same Latin script as English, which presents a different set of challenges.

We tested the traditional method, as well as a new approach where we included the top n words (not tokens) from the Indonesian vocabulary. We discovered that with the new approach, we could achieve better fertility scores by adding around 3000 new vocabulary words. Adding more than 3000 words did not significantly improve the fertility score further, but it increased the size of the embedding matrix, leading to longer training times.


\subsection{Optimizing for Efficiency} In our experiment of optimal performance, we ensured that the total number of tokens in our model's vocabulary is a multiple of 64. Drawing inspiration from the recent advancements in nanoGPT~\citep{KarpathyTweet}, we acknowledge the significance of seemingly minor details, such as this, in improving the computational efficiency. This strategic decision enhances the model's execution speed, allowing it to leverage computing power more effectively. The final iteration of our model produces an increased curated vocabulary consisting of 35,008 tokens, attuned to strike a harmonious balance between linguistic richness and computational efficiency.

Since we are expanding the tokenizer vocabulary, we also need to initialize the embeddings for these new vocabs. Building on the ideas presented in~\citep{hewitt2021initializing}, we initialize all new embeddings by averaging all existing embeddings. This approach limits the KL-divergence between the token-level distributions of language models before and after expansion, leading to improved performance during fine-tuning.

\subsection{Training \& Finetuning} For pretraining, our tokenizer processed about 8.79 billion tokens. Incremental pretraining, building upon Llama-2-7B-Base, was conducted over 3 epochs using LORA~\citep{hu2021lora}. This approach safeguards against catastrophic forgetting and optimizes hardware and cost requirements. The training utilized 8 x A100 40GB GPUs, taking approximately 300 hours. Supervised Fine-Tuning (SFT) further refined our model on diverse tasks for 5 epochs using LORA. Employing the same GPU configuration, the SFT process took about 36 hours. 

\section{Evaluation \& Results}

\subsection{Tokenizer Fertility Analysis}

\begin{table}[!b]
\centering
\small
\begin{tabular}{@{}cccccccc@{}}
\toprule
\multirow{2}{*}{\textbf{Model Name}} & \multicolumn{3}{c}{\textbf{Mean Fertility Score}}          & \multirow{2}{*}{\textbf{Vocab Size}} & \multicolumn{3}{c}{\textbf{\% Improvement}}                 \\ \cmidrule(lr){2-4} \cmidrule(l){6-8} 
                                     & \textbf{Indonesian} & \textbf{Regional} & \textbf{English} &                                      & \textbf{Indonesian} & \textbf{Regional} & \textbf{English} \\ \cmidrule(r){1-1} \cmidrule(lr){5-5}
Llama-2-7B                           & 2.858               & 2.658             & 1.666            & 32000                                & --                  & --                & --               \\
Komodo-7B                            & 2.031               & 1.996             & 1.633            & 35008                                & 28.90\%              & 24.90\%            & 1.98\%              \\ \bottomrule
\end{tabular}
\caption{Fertility Score Analysis. We compare the mean fertility scores of Llama-2-7B and Komodo-7B, measuring their token-splitting behavior across Indonesian, regional, and English languages.}
\label{tab:fertility}
\end{table}

In Natural Learning Processing, fertility refers to the ability of a tokenizer to break down text into meaningful units (tokens) while minimizing the number of unique tokens required. A highly fertile tokenizer will create many tokens from the original text, often by splitting words into smaller sub-units (like morphemes) or generating unique representations for unseen words. Conversely, a less fertile tokenizer will create fewer tokens, typically by relying on a vocabulary of known words.

In comparing the tokenizer performance between Llama-2-7B, our baseline model, and Komodo-7B, the enhanced version, notable distinctions emerge as shown in Table \ref{tab:fertility}. Llama-2-7B showcases mean fertility scores of 2.858 for Indonesian, 2.658 for regional languages, and 1.666 for English, with a vocabulary size of 32,000. On the other hand, Komodo-7B exhibits substantial improvements with mean fertility scores of 2.031 for Indonesian, 1.996 for regional languages, and 1.633 for English, coupled with an expanded vocabulary size of 35,008. This expansion results in significant percentage improvements—28.90\% for Indonesian, 24.90\% for regional languages, and 2\% for English—highlighting the improved tokenization capabilities of Komodo-7B compared to the Llama-2-7B baseline. This means during inference, Komodo-7B  will provide better latency compared to Llama-2. These results underscore the tangible advancements achieved through our model's refined tokenizer, offering improved word tokenization across diverse languages.

\begin{figure}[!b]
     \centering
     \includegraphics[width=0.90\textwidth]{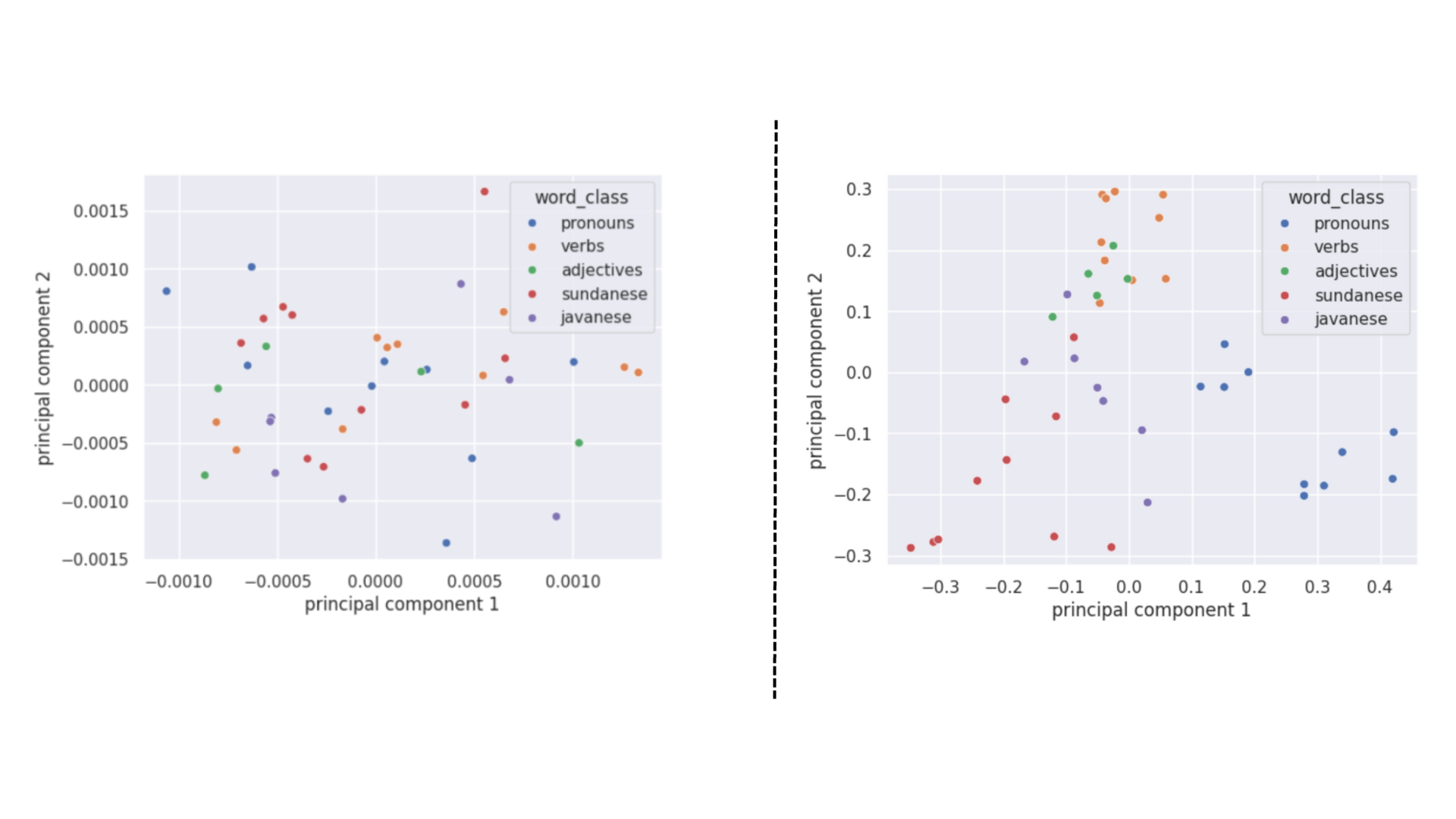}
     \caption{The left plot represents the initial embedding position of words when they are first randomly initialized, while the right plot shows their updated positions after 3 epochs of pre-training. The noticeable grouping of words from the same class in the right plot indicates effective learning and organization of word relationships during pre-training. These plots are created by utilizing PCA with 2 principal components.}
     \label{fig:embedding-position}
\end{figure}

\subsection{Embedding Position Analysis}

We aimed to examine the effectiveness of our model in refining embeddings over the course of pretraining. To do this, we selected 8-10 complete words from various word categories, beginning with pronouns, verbs, and adjectives, and progressing to include specific regional language words like Sundanese and Javanese. Initially, we plotted the embeddings' starting positions. Subsequently, we conducted three epochs of training and generated another plot to visualize the changes in the embeddings.

Figure \ref{fig:embedding-position} illustrates the initial and final positions of words in two plots by employing Principal Component Analysis to improve the interpretability of these plots. The plot on the left represents the initial positions, while the plot on the right shows their positions after three epochs of training. An important observation is that words belonging to the same group tend to cluster together after training. The results indicate an improvement in the model's ability to position similar words closer to each other in its memory as training progresses.

\subsection{Downstream Tasks}

This section evaluates Komodo-7B-Instruct performance across both discriminative and generative tasks. A comparative analysis with other models, including colloquial examples and English proficiency regression, enhances the comprehensive exploration of Komodo-7B-Instruct capabilities. There were situations during evaluation where the models produced results which couldn't be directly be evaluated using heuristics, so we used the help of GPT-4-0613 in that case, we have mentioned the prompts used for each task in the Appendix-section. As shown in Table \ref{tab:all-tasks}, Komodo-7B-Instruct outperforms various models across different tasks.

\subsubsection{Baselines}
We evaluate against multiple open-source and closed-source massively multilingual models to ensure a comprehensive evaluations.
\begin{itemize}
\item \textbf{GPT-3.5-turbo-0301}: Snapshot of GPT-3.5-turbo from March 1st 2023
\item \textbf{GPT-3.5-turbo-0613}: Snapshot of GPT-3.5-turbo from June 13th 2023
\item \textbf{GPT-3.5-turbo-1106}: GPT-3.5 Turbo model with improved instruction following, JSON mode, reproducible outputs, parallel function calling, and more. Returns a maximum of 4,096 output tokens.
\item \textbf{GPT-4-1106-preview}: Latest and best performing model from OpenAI giving State-of-the-art performance in many benchmarks till the date of benchmarking.
\item \textbf{Llama-2-7B-Chat}: The Llama2-7B-chat is a 7 billion parameter model from Meta and is trained on a large corpus of text that includes conversational data, such as chat logs and social media post.
\item \textbf{Llama-2-13B-Chat}: It is following same architecture as of it's 7B version, but with more parameters that is 13B.
\item \textbf{Mixtral-8x7B-v0.1-Instruct}: It is a pretrained generative Sparse Mixture of Experts from Mistral-AI. Each of these experts are specialized to optimize decisions for different tasks, the model uses a router to selectively activate a small subset of experts for each input, making it efficient to scale.

\item \textbf{Gemma-7B-it}~\citep{gemma}: It is  lightweight, instruction-tuned, 7-Billion parameter open model from Google, built from the same research and technology used to create the Gemini models. It is a text-to-text, decoder-only large language model.

\item \textbf{Sealion-7B-Instruct-NC}: SEA-LION is built on the robust MPT architecture by AI-Singapore and has a vocabulary size of 256K. The base model has been pre-trained specifically only on Southeast Asia (SEA) region languages. This model specifically is instruction-tuned only on Indonesian language data.

\item \textbf{Aya-101 (13B) }: The Aya model is a massively multilingual generative language model that follows instructions in 101 languages built by Cohere. This is an encoder-decoder instruction-tuned model built on top of mT5 base model. The Aya model is trained using xP3x, Aya Dataset, Aya Collection, a subset of DataProvenance collection and ShareGPT-Command.

\item \textbf{Bactrian-X-Llama-7B}: Bactrian is a LlaMA model finetuned on the Bactrian-X dataset which contains 3.4M pairs of instructions and responses in 52 languages. This dataset was automatically constructed by translating the Alpaca and Dolly Datasets using the Google Translate API.

\item \textbf{Qwen-1.5-7B-Chat}: Qwen1.5 is a language model series including decoder language models of different model sizes from Alibaba. It is based on the Transformer architecture with SwiGLU activation, attention QKV bias, group query attention, mixture of sliding window attention and full attention, etc. This model is the chat version of the model with 7B parameters.
\end{itemize}

\subsubsection{Discriminative Tasks}

Let's take a closer look at how Komodo-7B-Instruct is doing in different tasks. Table \ref{tab:all-tasks} provides insights that tell us how well Komodo-7B-Instruct understands and discriminates between various types of language challenges. For tasks like IndoMMLU, ID-EN (Indonesian-English) sentences 
\noindent
\begin{table}[!b]
\resizebox{\textwidth}{!}{%
\begin{tabular}{@{}
>{\columncolor[HTML]{FFFFFF}}c 
>{\columncolor[HTML]{FFFFFF}}l 
>{\columncolor[HTML]{FFFFFF}}c 
>{\columncolor[HTML]{FFFFFF}}c 
>{\columncolor[HTML]{FFFFFF}}c 
>{\columncolor[HTML]{FFFFFF}}c 
>{\columncolor[HTML]{FFFFFF}}c 
>{\columncolor[HTML]{FFFFFF}}c 
>{\columncolor[HTML]{FFFFFF}}c
>{\columncolor[HTML]{FFFFFF}}c
>{\columncolor[HTML]{FFFFFF}}c
>{\columncolor[HTML]{FFFFFF}}c @{}}
\toprule
\multirow{2}{*}{\textbf{Organization}} 
& \multirow{2}{*}{\textbf{Model Name}} 
& \multicolumn{7}{c}{\textbf{Discriminative Tasks}}
& \multicolumn{2}{c}{\textbf{Generative Tasks}}
& \multirow{2}{*}{\textbf{Average}} 
\\ \cmidrule(lr){3-9} \cmidrule(l){10-11} 
& \multicolumn{1}{p{1.5cm}}{} 
& \multicolumn{1}{p{1.5cm}}{\centering \textbf{Indo MMLU}} 
& \multicolumn{1}{p{1cm}}{\centering \textbf{ID-EN}} 
& \multicolumn{1}{p{1.5cm}}{\centering \textbf{XCOPA-ID}} 
& \multicolumn{1}{p{2.7cm}}{\centering \textbf{Intent Classification}} 
& \multicolumn{1}{p{2cm}}{\centering \textbf{Colloquial Detection}} 
& \multicolumn{1}{p{1.5cm}}{\centering \textbf{NusaX-Senti}} 
& \multicolumn{1}{p{1.75cm}}{\centering \textbf{ID-Hate Speech}} 
& \multicolumn{1}{p{1.5cm}}{\centering \textbf{TydiQA-ID}} 
& \multicolumn{1}{p{1.5cm}}{\centering \textbf{Indosum}} \\\midrule
\cellcolor[HTML]{FFFFFF}                                  & GPT-3.5-turbo-0301                                              & 51.3             & 64.5          & 70.0              & 82.0                          & 64.1                         & 47.2                       & 68.0                  & 85.3    &    41.0    &    63.7   \\
\cellcolor[HTML]{FFFFFF}                                  & GPT-3.5-turbo-0613                                              & 52.7             & 66.8          & 88.2              & 84.0                          & 75.1                         & 63.3                       & 63.7                  & 86.4    &    40.0       &    68.9     \\
\cellcolor[HTML]{FFFFFF}                                  & GPT-3.5-turbo-1106                                              & 53.3             & 69.7          & 89.3              & 84.0                          & 64.2                         & 59.8                       & 56.6                  & 88.0     &   42.0      &    67.4      \\
\multirow{-4}{*}{\cellcolor[HTML]{FFFFFF}\textbf{OpenAI}} & GPT-4-preview-1106                                              & \textbf{69.8}    & 78.0          & \textbf{98.3}     & \textbf{89.0}                 & \textbf{92.7}                & 66.1                       & \textbf{73.4}         & 72.0      & 33.0  &    \textbf{74.7} \\ \hline
\cellcolor[HTML]{FFFFFF}                                  & Llama-2-7B-Chat                                                 & 30.4             & 45.6          & 41.5              & 57.0                          & 31.4                         & 2.9                        & 41.3                  & 11.7       & 34.0          &    32.9  \\
\multirow{-2}{*}{\cellcolor[HTML]{FFFFFF}\textbf{Meta}}   & Llama-2-13B-Chat                                                & 32.0             & 61.7          & 38.0              & 59.0                          & 31.1                         & 58.7                       & 57.2                  & 71.9       & 40.0      &    50.0      \\ \hline
\textbf{Google}                                           & Gemma-7B-it                                                     & 37.4             & 73.6         & 57.7              & 77.1                          & 18.8                         & 44.2                       & 54.8                  & 73.3       & 44.0       &    53.4     \\ \hline
\textbf{Mistral}                                          & Mixtral-8x7B-v0.1-Instruct                                      & 45.2             & 57.8          & 88.7              & 86.0                          & 41.1                         & 52.8                       & 68.8                  & 90.3       &   14.0       &    60.5     \\ \hline
\textbf{AISingapore}                                      & Sealion-7B-Instruct-NC                                          & 23.9             & 26.9          & 41.3              & 37.0                          & 41.8                         & 30.7                       & 57.3                  & 65.3        &  26.0      &    38.9      \\ \hline
\textbf{Cohere}                                           & Aya-101-13B                                                     & 47.7             & 47.3          & 84.0              & 64.0                          & 18.9                         & 74.6                       & 72.7                  & 81.3        & 39.0       &    58.8     \\ \hline
\textbf{MBZUAI}                                           & Bactrian-X-Llama-7B                                             & 23.6             & 43.2          & 45.3              & 42.0                          & 50.3                         & 44.5                       & 42.4                  & 65.0        & 15.0        &    41.3    \\ \hline
\textbf{Alibaba}                                             & Qwen-1.5-7B-chat                                                & 40.0             & 56.0          & 29.5              & 85.0                          & 41.8                         & 58.7                       & 63.9                  & 51.22       & 29.0       &    50.6     \\ \hline
\textbf{Yellow.ai}                                        & Komodo-7B-Instruct                                                       & 43.2             & \textbf{90.5} & 79.6              & 84.0                          & 73.6                         & \textbf{79.3}              & 56.2                  & \textbf{90.3}  & \textbf{43.0}   &    \textbf{71.1}\\ \bottomrule
\end{tabular}
}
\caption{This table breaks down how well Komodo-7B-Instruct tackles various language tasks compared to other models. \textbf{Notes:} (1) For Sealion \& Mixtral, we have used the prompts provided by the authors. (2) Performance of GPT4  in the TydiQA-ID data is low because the model refuse to answer the query most of the time due to hallucination prevention. (3) All evaluation functions are attached in the Appendix section.}
\label{tab:all-tasks}
\end{table} , and X-Copa-ID, Komodo-7B-Instruct consistently scores well, showing it's good at understanding different kinds of language tasks. Special focus is its high score of 90.5 in ID-EN, which is better than many other models. This shows that Komodo-7B-Instruct excels in cross-language understanding, especially between Indonesian and English.

In tasks like Intent Classification, where it figures out what users are trying to do, and Colloquial Detection, where it understands informal language, Komodo-7B-Instruct does really well. Scoring above 80 in Intent Classification and excelling in Colloquial Detection shows that Komodo-7B-Instruct is versatile—it gets what people mean and can handle casual language.

When it comes to understanding sentiment in different languages, Komodo-7B-Instruct has shown it's really good. We can see that Komodo-7B-Instruct achieves SOTA on the NusaX-Senti dataset, with a spectacular score of 79.3.

\begin{figure*}[!t]
     \centering
     \includegraphics[width=1\textwidth]{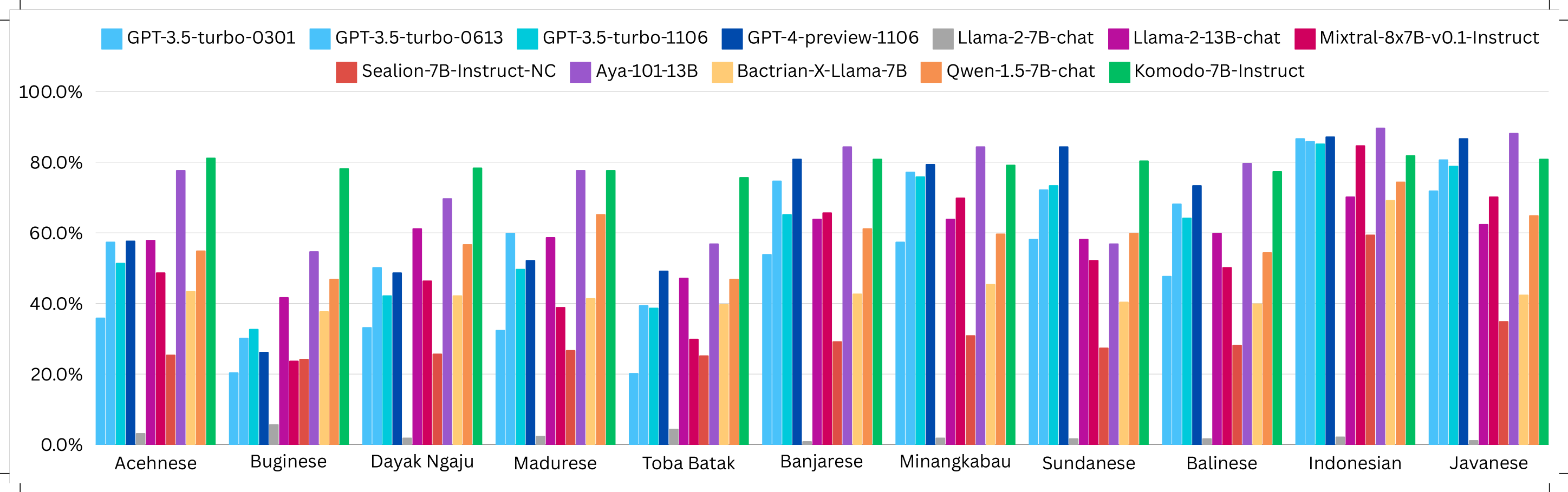}
     \caption{Performance breakdown of all models on NusaX-Senti dataset.}
     \label{fig:language}
\end{figure*}

Figure \ref{fig:language} further analyze the performance breakdown of all benchmarked models on each of the 11 languages within the NusaX-Senti dataset. We can see that Komodo-7B-Instruct excels especially in languages like Acehnese, Buginese, Dayak Ngaju, Madurese, and Toba Batak. In these languages, Komodo-7B-Instruct is the top-performer, being the best at figuring out if the words express positive or negative sentiment.

In other languages like Banjarese, Minangkabau, Sundanese, and Balinese, Komodo-7B-Instruct stands second, just behind models including GPT-4-1106-preview and Aya-101-13B. One of our future work includes training Komodo-13B-Instruct, which we expects to outperform other models in all languages

\subsubsection{Generative Tasks}

Table \ref{tab:all-tasks} also shows how Komodo-7B-Instruct performs in generative tasks. In the field of generative tasks, TydiQA-ID assess the model's ability to answer questions, while Indosum focuses on summarization creation. Komodo-7B-Instruct stands out prominently, achieving a remarkable accuracy of 90.3 in TydiQA-ID, indicating its proficiency in generating relevant and informative answers. Additionally, in the Indosum task, which involves summarizing content, Komodo-7B-Instruct secures an exceptional score of 43.0, showcasing its capability in condensing information coherently. Comparison with other models like GPT-3.5, Llama-2-Chat, Mixtral-8X7B, and Aya, reveal Komodo-7B-Instruct consistent performance and effectiveness in generative language tasks.

\begin{figure*}[!t]
     \centering
     \includegraphics[width=0.7\textwidth]{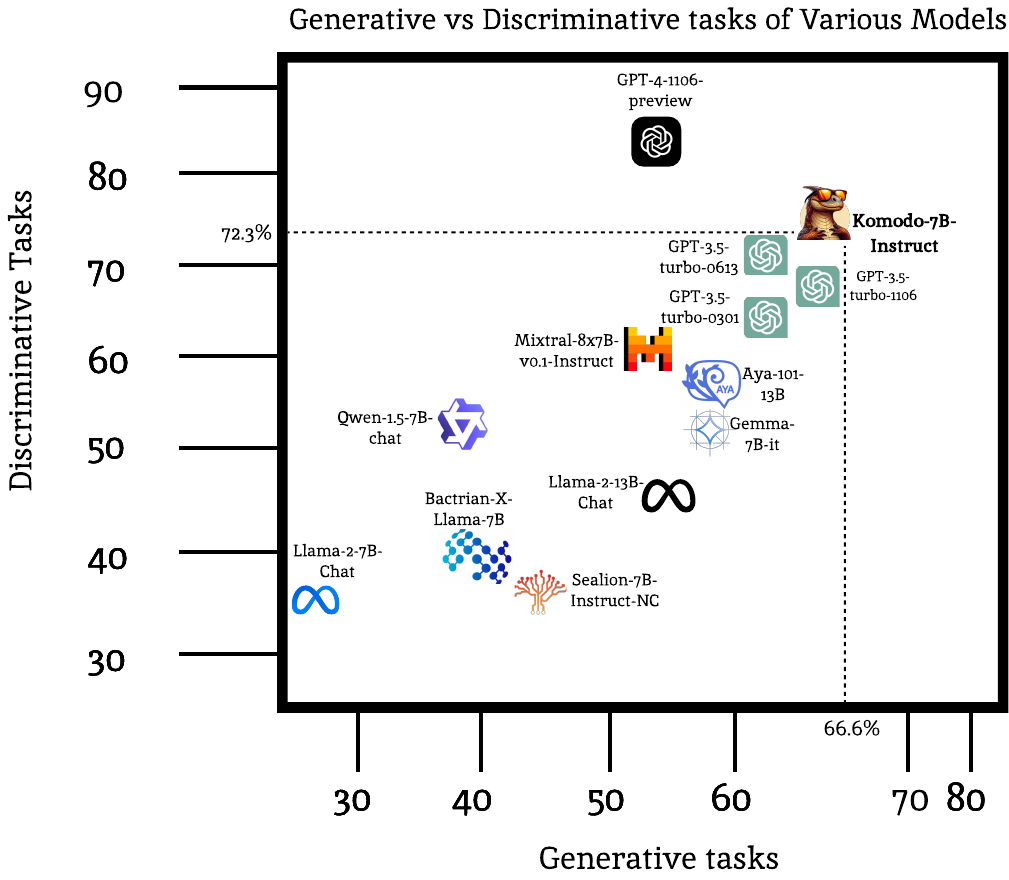}
     \caption{A plot illustrating Komodo-7B-Instruct adeptness in balancing generative and discriminative tasks, showcasing strong performance across diverse language challenges.}
     \label{fig:scatter_plot}
\end{figure*}

\subsubsection{Generative vs Discriminative}

In Figure \ref{fig:scatter_plot}, the plot provides a visual comparison of Komodo-7B-Instruct performance in generative and discriminative tasks. Each point on the plot corresponds to a specific model, showcasing how well Komodo-7B-Instruct balances the creation of meaningful content and accurate discrimination across diverse language challenges. Komodo-7B-Instruct excels in both Generative and Discriminative tasks, scoring 66.6 and 72.3, respectively. It demonstrates strong proficiency, trailing only slightly behind the GPT-4-1106-preview model, which excels notably in Discriminative tasks with a score of 81. This shows Komodo-7B-Instruct robust performance and versatility in handling various language tasks.

\subsubsection{Translation Performance Comparison with Google Translate}

\begin{figure*}[!t]
     \centering
     \includegraphics[width=1\textwidth]{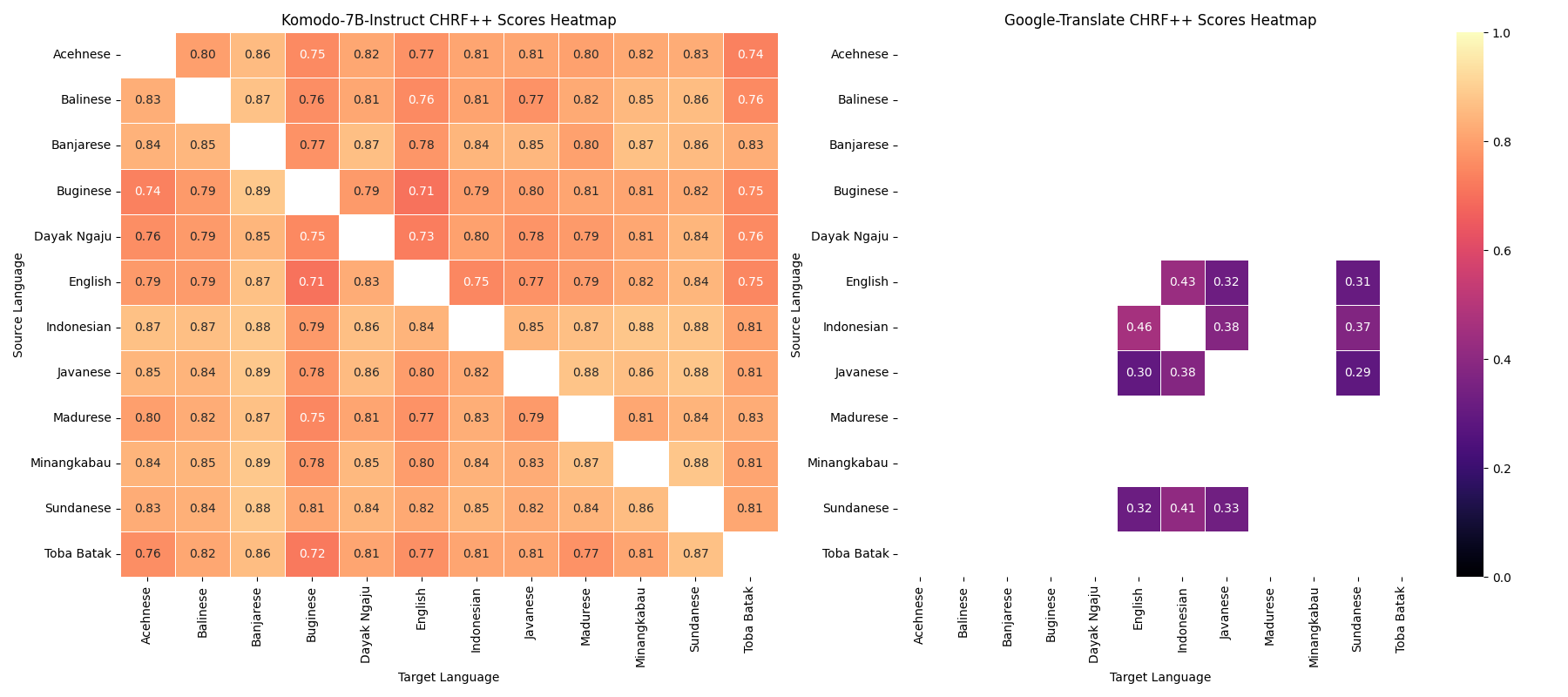}
     \caption{A comparison between the Google-translate \& Komodo-7B-Instruct}
     \label{fig:heatmap}
\end{figure*}

Figure \ref{fig:heatmap} serves as a valuable analysis for evaluating the translation capabilities of Komodo-7B-Instruct in comparison to Google Translate. The visual representation allows us to discern the languages each platform supports. On the right side, the heatmap illustrates Google Translate's proficiency, primarily in Javanese, English, Indonesian, and Sundanese. However, this leaves numerous language spaces unoccupied.

Conversely, the left side of the heatmap showcases the comprehensive linguistic capabilities of Komodo-7B-Instruct, encompassing a total of 11 regional languages. This inclusive approach extends the reach of education in Indonesia by enabling direct translation from English to a diverse set of regional languages including languages which are not supported by many models and translation systems like Acehnese, Balinese, Banjarese, Buginese, Madurese, Minangkabau, and Toba Batak. Our benchmarking data doesn't contain Lampungnese but it is worth noting that Komodo-7B-Instruct also supports Lampungnese. This is a significant advancement, considering that Google Translate's support is limited to Javanese and Sundanese, primarily spoken in Java. Additionaly, Komodo-7B-Instruct can translate between the regional languages without the need of any intermediate language such as English or Indonesian.

The broader coverage of Komodo-7B-Instruct ensures that individuals across various regions in Indonesia, beyond Java, can benefit from education in their native languages. This not only enhances accessibility but also addresses the challenge of language diversity in educational settings. Therefore, Komodo-7B-Instruct stands as a promising solution for bridging educational gaps and encouraging inclusivity in language learning.

\subsubsection{Comparison with Other Fine-Tuned Models}

\begin{figure*}[!t]
     \centering
     \includegraphics[width=0.8\textwidth]{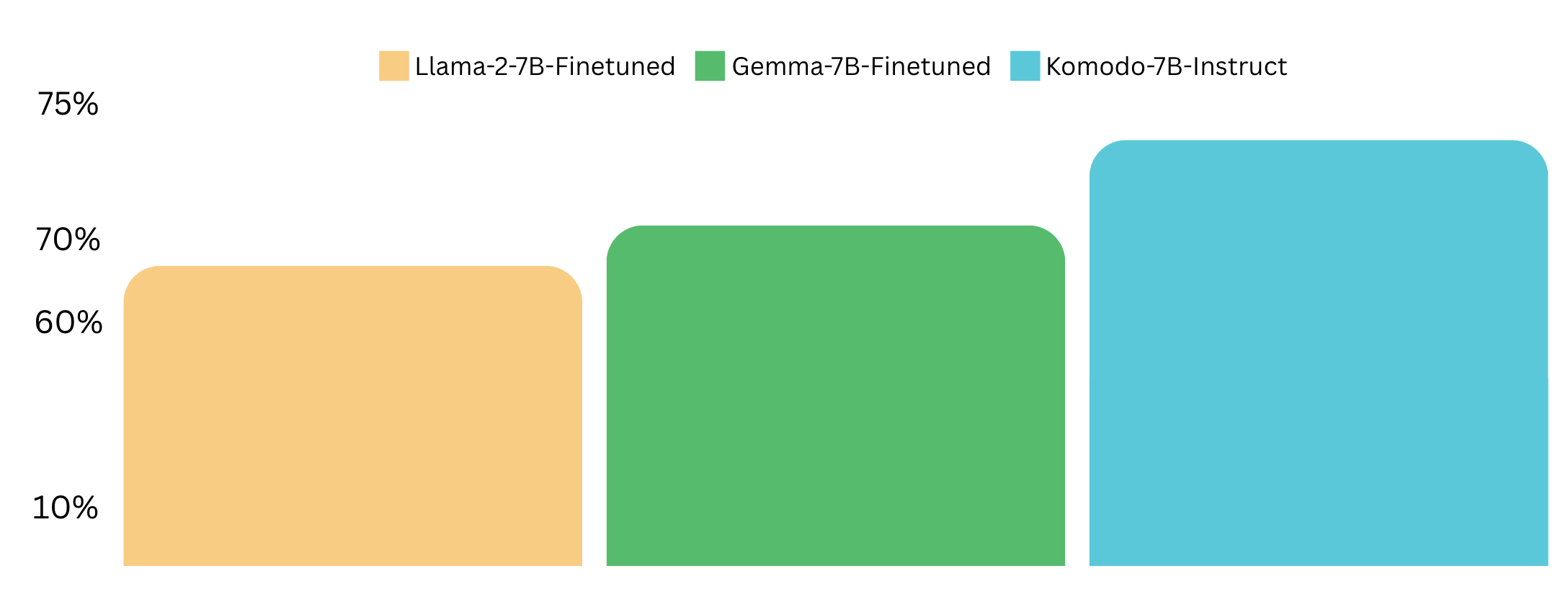}
     \caption{Performance Comparison: Komodo-7B-Instruct versus Llama-2-7B-Finetuned versus Gemma-7B-Finetuned across diverse language tasks}
     \label{fig:komodo-competitors}
\end{figure*}

Figure~\ref{fig:komodo-competitors} shows that Komodo-7B-Instruct demonstrates robust performance across both generative and discriminative tasks, achieving an average score of 72.63\%. In comparison, Gemma-7B-finetuned scores slightly lower at 70.1\%, and  Llama-2-7B-finetuned stands at 68.9\%.

It is essential to highlight that we manually fine-tuned Llama-2-7B-Base and Gemma-7B-base using the same SFT data employed to train Komodo-7B-Instruct. This approach aims to demonstrate that the pretraining phase conducted on Komodo-7B-Instruct contributes significantly to the enhancement of language understanding capabilities for Indonesian and regional languages. The incremental pretraining performed on the Llama-2-7B-Base model proves beneficial, enabling Komodo-7B-Instruct to achieve superior performance compared to Llama-2-7B-Finetuned.It is worth noting that Gemma shows promising results as a base model compared to Llama-2.

\subsubsection{Qualitative Analysis}

\begin{figure}[!b]
     \centering
     \includegraphics[width=1\textwidth]{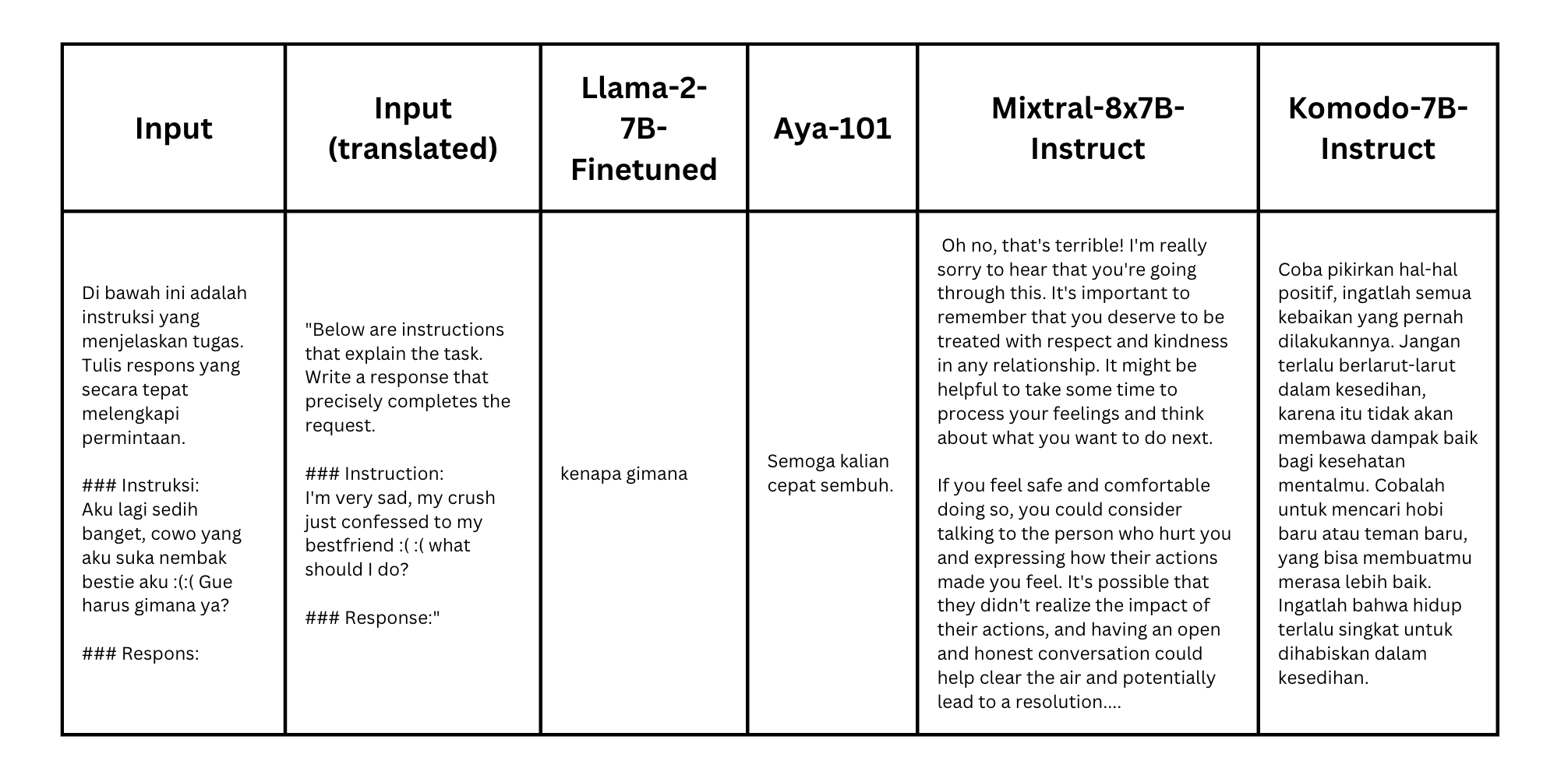}
     \caption{A sample of general instruction and responses from each model. "Input (translated)" is the translated version of the original "Input" that is passed to the model, this is only for visualization and understanding purpose for non-Indonesian readers.Additional samples of general instructions can be found in the Appendix.}
     \label{fig:manual_instructions}
\end{figure}

In addition to quantitative benchmarking, we conducted qualitative testing by providing various general instructions to the model. Figure \ref{fig:manual_instructions} displays a sample of these instructions along with the responses from Llama-2-7B-Finetuned, Aya-101, Mixtral-8x7B-Instruct, and Komodo-7B-Instruct. Notably, the response from Mixtral-8x7B-Instruct is in English, despite the input being in Indonesian. While Llama-2-7b-Finetuned responds in Indonesian, the reply lacks empathy and helpfulness. Aya-101's response is more empathetic but falls short in terms of helpfulness. Conversely, Komodo-7B-Instruct response is highly empathetic and helpful, offering both consolation and solutions.

\begin{table}[!t]
\centering
\footnotesize
\begin{tabular}{@{}lcc@{}}
\toprule
\textbf{Model} & \textbf{English PPL} & \textbf{Indonesian PPL} \\ \midrule
Llama-2-7B-Base     & 7.7173               & 6.2608              \\
Komodo-7B-Base      & 7.8648               & 5.6477              \\ \bottomrule
\end{tabular}
\caption{This illustrates perplexity benchmarking results for Llama-2-7B \& Komodo-7B, emphasizing Komodo-7B's competitive performance in both English and Indonesian contexts.}
\label{table-perplexity}
\end{table}

\subsubsection{English Capability Regression Analysis}

Perplexity serves as a significant metric for evaluating pretrained language models. Table \ref{table-perplexity} presents the perplexity benchmarking results for Llama-2-7B-Base and Komodo-7B-Base, focusing on their performance in both English and Indonesian contexts. Llama-2-7B-Base demonstrates a perplexity of 7.7173 in English and 6.2608 in Indonesian, while Komodo-7B-Base exhibits a slightly higher perplexity of 7.8648 in English and a lower perplexity of 5.6477 in Indonesians. These figures provide valuable insights into the language comprehension capabilities of the models, with lower perplexity values indicating enhanced predictive performance. The analysis underscores Komodo-7B-Base's competitive performance, maintaining favorable perplexity levels in both English and Indonesian language domains.

\begin{figure*}[!b]
     \centering
     \includegraphics[width=1\textwidth]{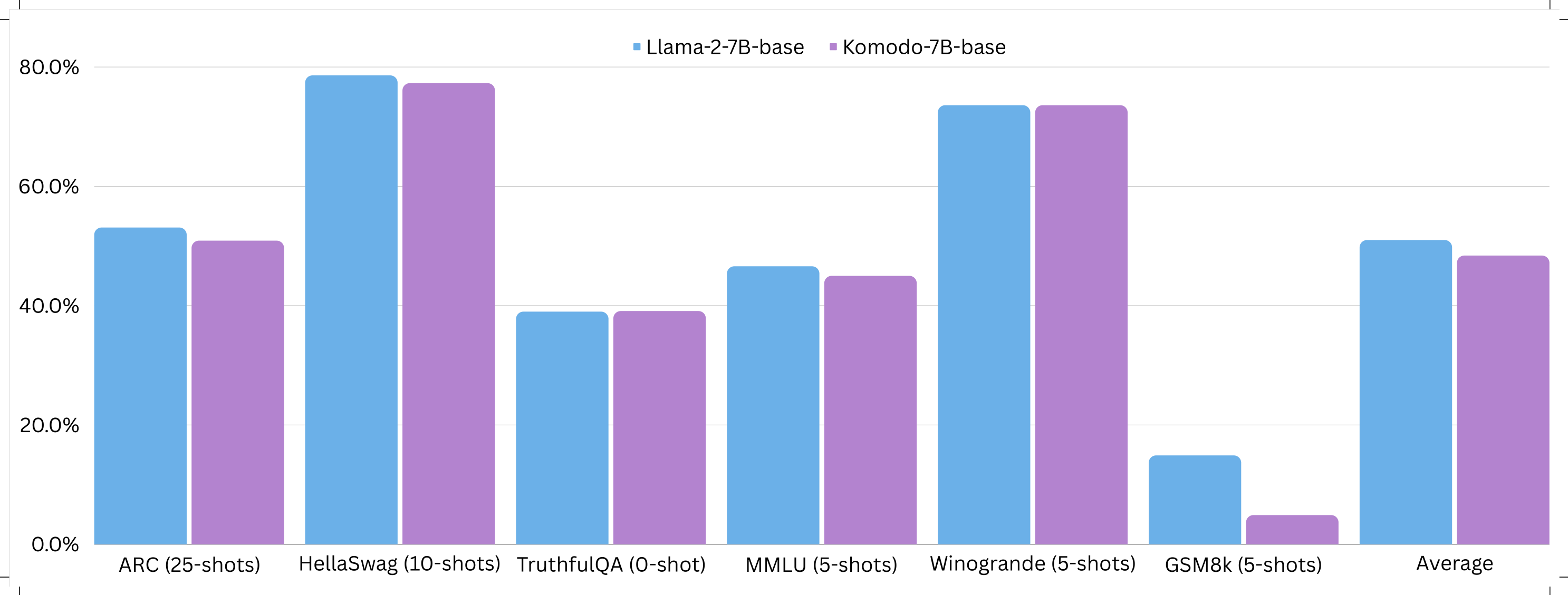}
     \caption{Comparative Analysis of Komodo-7B-Base and Llama-2-7B-Base on English Language Tasks}
     \label{fig:english-benchmarks}
\end{figure*}

In a detailed examination of Komodo-7B-Base and Llama-2-7B-Base performances across various English language downstream tasks, Komodo-7B-Base consistently demonstrates proficiency comparable to Llama-2-7B-Base in several instances, affirming its competence in handling diverse linguistic challenges. Notably, Komodo-7B-Base achieves competitive scores in tasks such as ARC (25-shots), HellaSwag (10-shots), TruthfulQA (0-shot), MMLU (5-shots), Winogrande (5-shots), and GSM8k (5-shots), showcasing its versatility and linguistic prowess. We utilized the LM Evaluation Harness~\citep{eval-harness} repository to perform the benchmarking on these datasets.

As shown in the Figure \ref{fig:english-benchmarks}, Komodo-7B-Base is able to maintain the performance of Llama-2-7B-Base across all tasks, except GSM8k, which consists of mathematical task. This probably happens because our pre-training data consists of very less mathematical data. However, it is crucial to underscore that Komodo-7B-Base primary focus lies in its exceptional proficiency in Indonesian and  Indonesia's regional languages. The model's design and fine-tuning prioritize linguistic variations specific to the Indonesian context and it's regional languages, enabling it to outperform in tasks related to Indonesian and regional languages. Thus, any minor disparities in English tasks should be viewed in the context of Komodo-7B-Base specialized strength in catering to the linguistic diversity of Indonesia, showcasing its effectiveness in its designated domain.

\section{Conclusion and Future Works}

\subsection{Conclusion}

In this paper, we introduced Komodo-7B, a robust language model tailored for Indonesian and regional languages. Our extensive evaluation demonstrated its impressive performance across various tasks, showcasing versatility in both generative and discriminative challenges. The model's balanced proficiency makes it well-suited for addressing diverse language nuances. The collaborative efforts and advancements in architecture and training techniques have resulted in a valuable resource for linguistically rich regions. Furthermore, it's noteworthy that Komodo-7B-Instruct supports 11 Indonesia's regional languages not covered by Google Translate. Models accommodating these languages often exhibit subpar performance, emphasizing Komodo-7B-Instruct importance in bridging the language gap for these linguistic nuances.

\subsection{Future Works}

Looking ahead, our future work involves the development and exploration of bigger models, such as the 13B parameters variant of the model. The aim is to push the boundaries of language understanding and generation, providing enhanced capabilities for addressing complex linguistic nuances and challenges. The journey towards linguistic excellence continues, and we are enthusiastic about the possibilities that a future iteration of Komodo can bring to the field of multilingual language models. Our commitment to contributing to advancements in this domain remains steadfast.

\section{Acknowledgement}
We want to express our heartfelt thanks to the amazing teams at vLLM~\citep{vllm} and Text Generation Inference~\citep{tgi} (by Hugging Face). Their tools have been a huge help in our research journey. With these tools, we managed to boost our model's data generation by multiple times! We owe a big part of our success to the fantastic support and resources provided by these teams. The ability to distribute parallel requests across multiple GPUs using tensor parallelism has made our work much more efficient. This acknowledgment is a shout-out to the collaborative spirit and the wonderful possibilities that open-source communities bring. We are truly grateful.

\bibliography{addon}

\begin{thebibliography}{38}
\providecommand{\natexlab}[1]{#1}
\providecommand{\url}[1]{\texttt{#1}}
\expandafter\ifx\csname urlstyle\endcsname\relax
  \providecommand{\doi}[1]{doi: #1}\else
  \providecommand{\doi}{doi: \begingroup \urlstyle{rm}\Url}\fi

\bibitem[AISingapore(2023)]{sea_lion_2023}
AISingapore.
\newblock Sea-lion (southeast asian languages in one network): A family of large language models for southeast asia.
\newblock \url{https://github.com/aisingapore/sealion}, 2023.

\bibitem[Alfina et~al.(2017)Alfina, Mulia, Fanany, and Ekanata]{idhatespeech}
Ika Alfina, Rio Mulia, {Mohamad Ivan} Fanany, and Yudo Ekanata.
\newblock Hate speech detection in the indonesian language: A dataset and preliminary study.
\newblock In \emph{2017 International Conference on Advanced Computer Science and Information Systems, ICACSIS 2017}, pp.\  233--237. Institute of Electrical and Electronics Engineers Inc., 2017.
\newblock \doi{10.1109/ICACSIS.2017.8355039}.

\bibitem[Bai et~al.(2024)Bai, Bai, Chu, Cui, Dang, Deng, Fan, Ge, Huang, Hui, Li, Lin, Lin, Liu, Liu, Lu, Ma, Men, Ni, Ren, Ren, San, Tan, Tu, Wang, Wang, Xu, Yang, Yang, Yang, Yang, Yao, Yu, Zhang, Zhang, Zhang, Zheng, Zhou, Zhou, Zhou, and Zhu]{qwen15}
Jinze Bai, Shuai Bai, Yunfei Chu, Zeyu Cui, Kai Dang, Xiaodong Deng, Yang Fan, Wenbin Ge, Fei Huang, Binyuan Hui, Mei Li, Junyang Lin, Runji Lin, Dayiheng Liu, Tianyu Liu, Keming Lu, Jianxin Ma, Rui Men, Na~Ni, Xingzhang Ren, Xuancheng Ren, Zhou San, Sinan Tan, Jianhong Tu, Peng Wang, Shijie Wang, Jin Xu, An~Yang, Jian Yang, Kexin Yang, Shusheng Yang, Yang Yao, Bowen Yu, Jianwei Zhang, Yichang Zhang, Zhenru Zhang, Bo~Zheng, Chang Zhou, Jingren Zhou, Xiaohuan Zhou, and Tianhang Zhu.
\newblock Qwen1.5.
\newblock Technical report, Alibaba, 2024.
\newblock URL \url{https://qwenlm.github.io/blog/qwen1.5/}.

\bibitem[Brown et~al.(2020)Brown, Mann, Ryder, Subbiah, Kaplan, Dhariwal, Neelakantan, Shyam, Sastry, Askell, Agarwal, Herbert-Voss, Krueger, Henighan, Child, Ramesh, Ziegler, Wu, Winter, Hesse, Chen, Sigler, Litwin, Gray, Chess, Clark, Berner, McCandlish, Radford, Sutskever, and Amodei]{Chat-GPT}
Tom~B. Brown, Benjamin Mann, Nick Ryder, Melanie Subbiah, Jared Kaplan, Prafulla Dhariwal, Arvind Neelakantan, Pranav Shyam, Girish Sastry, Amanda Askell, Sandhini Agarwal, Ariel Herbert-Voss, Gretchen Krueger, Tom Henighan, Rewon Child, Aditya Ramesh, Daniel~M. Ziegler, Jeffrey Wu, Clemens Winter, Christopher Hesse, Mark Chen, Eric Sigler, Mateusz Litwin, Scott Gray, Benjamin Chess, Jack Clark, Christopher Berner, Sam McCandlish, Alec Radford, Ilya Sutskever, and Dario Amodei.
\newblock Language models are few-shot learners.
\newblock In \emph{Advances in Neural Information Processing Systems}, volume~33, pp.\  1877--1901, 2020.
\newblock URL \url{https://proceedings.neurips.cc/paper_files/paper/2020/file/1457c0d6bfcb4967418bfb8ac142f64a-Paper.pdf}.

\bibitem[Cahyawijaya et~al.(2021)Cahyawijaya, Winata, Wilie, Vincentio, Li, Kuncoro, Ruder, Lim, Bahar, Khodra, Purwarianti, and Fung]{cahyawijaya-etal-2021-indonlg}
Samuel Cahyawijaya, Genta~Indra Winata, Bryan Wilie, Karissa Vincentio, Xiaohong Li, Adhiguna Kuncoro, Sebastian Ruder, Zhi~Yuan Lim, Syafri Bahar, Masayu~Leylia Khodra, Ayu Purwarianti, and Pascale Fung.
\newblock Indonlg: Benchmark and resources for evaluating indonesian natural language generation.
\newblock In \emph{Proceedings of the 2021 Conference on Empirical Methods in Natural Language Processing}, pp.\  8875--8898, November 2021.
\newblock URL \url{https://aclanthology.org/2021.emnlp-main.699}.

\bibitem[Clark et~al.(2020)Clark, Choi, Collins, Garrette, Kwiatkowski, Nikolaev, and Palomaki]{clark-etal-2020-tydi}
Jonathan~H. Clark, Eunsol Choi, Michael Collins, Dan Garrette, Tom Kwiatkowski, Vitaly Nikolaev, and Jennimaria Palomaki.
\newblock Tydiqa a benchmark for information-seeking question answering in typologically diverse languages.
\newblock \emph{Transactions of the Association for Computational Linguistics}, 8:\penalty0 454--470, 2020.
\newblock URL \url{https://aclanthology.org/2020.tacl-1.30}.

\bibitem[Conover et~al.(2023)Conover, Hayes, Mathur, Xie, Wan, Shah, Ghodsi, Wendell, Zaharia, and Xin]{DatabricksBlog2023DollyV2}
Mike Conover, Matt Hayes, Ankit Mathur, Jianwei Xie, Jun Wan, Sam Shah, Ali Ghodsi, Patrick Wendell, Matei Zaharia, and Reynold Xin.
\newblock Free dolly: Introducing the world's first truly open instruction-tuned llm, 2023.
\newblock URL \url{https://www.databricks.com/blog/2023/04/12/dolly-first-open-commercially-viable-instruction-tuned-llm}.

\bibitem[Cui et~al.(2023)Cui, Yang, and Yao]{Chinese-LLaMA-Alpaca}
Yiming Cui, Ziqing Yang, and Xin Yao.
\newblock Efficient and effective text encoding for chinese llama and alpaca.
\newblock \emph{arXiv preprint arXiv:2304.08177}, 2023.
\newblock URL \url{https://arxiv.org/abs/2304.08177}.

\bibitem[DeepMind(2024)]{gemma}
Google DeepMind.
\newblock Gemma: Introducing new state-of-the-art open models.
\newblock Technical report, Google DeepMind, February 2024.
\newblock URL \url{https://storage.googleapis.com/deepmind-media/gemma/gemma-report.pdf}.

\bibitem[Ding et~al.(2023)Ding, Chen, Xu, Qin, Zheng, Hu, Liu, Sun, and Zhou]{ding2023enhancing}
Ning Ding, Yulin Chen, Bokai Xu, Yujia Qin, Zhi Zheng, Shengding Hu, Zhiyuan Liu, Maosong Sun, and Bowen Zhou.
\newblock Enhancing chat language models by scaling high-quality instructional conversations.
\newblock \emph{arXiv preprint arXiv:2305.14233}, 2023.
\newblock URL \url{https://arxiv.org/pdf/2305.14233.pdf}.

\bibitem[Face(2023)]{tgi}
Hugging Face.
\newblock Text generation inference, 2023.
\newblock URL \url{https://github.com/huggingface/text-generation-inference}.

\bibitem[Gunasekar et~al.(2023)Gunasekar, Zhang, Aneja, Mendes, Giorno, Gopi, Javaheripi, Kauffmann, de~Rosa, Saarikivi, Salim, Shah, Behl, Wang, Bubeck, Eldan, Kalai, Lee, and Li]{gunasekar2023textbooks}
Suriya Gunasekar, Yi~Zhang, Jyoti Aneja, Caio César~Teodoro Mendes, Allie~Del Giorno, Sivakanth Gopi, Mojan Javaheripi, Piero Kauffmann, Gustavo de~Rosa, Olli Saarikivi, Adil Salim, Shital Shah, Harkirat~Singh Behl, Xin Wang, Sébastien Bubeck, Ronen Eldan, Adam~Tauman Kalai, Yin~Tat Lee, and Yuanzhi Li.
\newblock Textbooks are all you need.
\newblock \emph{arXiv preprint arXiv:2306.11644}, 2023.
\newblock URL \url{https://arxiv.org/pdf/2306.11644.pdf}.

\bibitem[Hewitt(2021)]{hewitt2021initializing}
John Hewitt.
\newblock Initializing new word embeddings for pretrained language models, 2021.
\newblock URL \url{https://nlp.stanford.edu/~johnhew/vocab-expansion.html}.

\bibitem[Hu et~al.(2021)Hu, Shen, Wallis, Allen-Zhu, Li, Wang, Wang, and Chen]{hu2021lora}
Edward~J. Hu, Yelong Shen, Phillip Wallis, Zeyuan Allen-Zhu, Yuanzhi Li, Shean Wang, Lu~Wang, and Weizhu Chen.
\newblock Lora: Low-rank adaptation of large language models, 2021.
\newblock URL \url{https://arxiv.org/abs/2106.09685}.

\bibitem[Jiang et~al.(2024)Jiang, Sablayrolles, Roux, Mensch, Savary, Bamford, Chaplot, de~las Casas, Hanna, Bressand, Lengyel, Bour, Lample, Lavaud, Saulnier, Lachaux, Stock, Subramanian, Yang, Antoniak, Scao, Gervet, Lavril, Wang, Lacroix, and Sayed]{jiang2024mixtral}
Albert~Q. Jiang, Alexandre Sablayrolles, Antoine Roux, Arthur Mensch, Blanche Savary, Chris Bamford, Devendra~Singh Chaplot, Diego de~las Casas, Emma~Bou Hanna, Florian Bressand, Gianna Lengyel, Guillaume Bour, Guillaume Lample, Lélio~Renard Lavaud, Lucile Saulnier, Marie-Anne Lachaux, Pierre Stock, Sandeep Subramanian, Sophia Yang, Szymon Antoniak, Teven~Le Scao, Théophile Gervet, Thibaut Lavril, Thomas Wang, Timothée Lacroix, and William~El Sayed.
\newblock Mixtral of experts.
\newblock \emph{arXiv preprint arXiv:2401.04088}, 2024.
\newblock URL \url{https://arxiv.org/pdf/2401.04088.pdf}.

\bibitem[Karpathy(2023)]{KarpathyTweet}
Andrej Karpathy.
\newblock Tweet on nanogpt optimization, February 2023.
\newblock URL \url{https://twitter.com/karpathy/status/1621578354024677377}.

\bibitem[Koto et~al.(2023)Koto, Aisyah, Li, and Baldwin]{koto-etal-2023-indommlu}
Fajri Koto, Nurul Aisyah, Haonan Li, and Timothy Baldwin.
\newblock Large language models only pass primary school exams in indonesia: A comprehensive test on indommlu".
\newblock In \emph{Proceedings of the 2023 Conference on Empirical Methods in Natural Language Processing (EMNLP)}, December 2023.

\bibitem[Kurniawan \& Louvan(2018)Kurniawan and Louvan]{indosum}
Kemal Kurniawan and Samuel Louvan.
\newblock Indosum: A new benchmark dataset for indonesian text summarization.
\newblock In \emph{2018 International Conference on Asian Language Processing (IALP)}, pp.\  215--220, 2018.
\newblock \doi{10.1109/IALP.2018.8629109}.

\bibitem[Kwon et~al.(2023)Kwon, Li, Zhuang1, Sheng1, Lianmin~Zheng, Gonzalez, Zhang, and Stoica]{vllm}
Woosuk Kwon, Zhuohan Li, Siyuan Zhuang1, Ying Sheng1, Cody Hao~Yu Lianmin~Zheng, Joseph~E. Gonzalez, Hao Zhang, and Ion Stoica.
\newblock Attention is all you need.
\newblock In \emph{Proceedings of the 29th Symposium on Operating Systems Principles}, pp.\  611–626, October 2023.
\newblock URL \url{https://doi.org/10.1145/3600006.3613165}.

\bibitem[Köpf et~al.(2023)Köpf, Kilcher, von Rütte, Anagnostidis, Tam, Stevens, Barhoum, Duc, Stanley, Nagyfi, ES, Suri, Glushkov, Dantuluri, Maguire, Schuhmann, Nguyen, and Mattick]{köpf2023openassistant}
Andreas Köpf, Yannic Kilcher, Dimitri von Rütte, Sotiris Anagnostidis, Zhi-Rui Tam, Keith Stevens, Abdullah Barhoum, Nguyen~Minh Duc, Oliver Stanley, Richárd Nagyfi, Shahul ES, Sameer Suri, David Glushkov, Arnav Dantuluri, Andrew Maguire, Christoph Schuhmann, Huu Nguyen, and Alexander Mattick.
\newblock Openassistant conversations - democratizing large language model alignment.
\newblock \emph{arXiv preprint arXiv:2304.07327}, 2023.
\newblock URL \url{https://arxiv.org/pdf/2304.07327.pdf}.

\bibitem[Li et~al.(2023{\natexlab{a}})Li, Koto, Wu, Aji, and Baldwin]{li2023bactrianx}
Haonan Li, Fajri Koto, Minghao Wu, Alham~Fikri Aji, and Timothy Baldwin.
\newblock Bactrian-x: Multilingual replicable instruction-following models with low-rank adaptation.
\newblock \emph{arXiv preprint arXiv:2305.15011}, 2023{\natexlab{a}}.
\newblock URL \url{https://arxiv.org/pdf/2303.08774.pdf}.

\bibitem[Li et~al.(2023{\natexlab{b}})Li, Zhang, Dubois, Taori, Gulrajani, Guestrin, Liang, and Hashimoto]{alpaca_eval}
Xuechen Li, Tianyi Zhang, Yann Dubois, Rohan Taori, Ishaan Gulrajani, Carlos Guestrin, Percy Liang, and Tatsunori~B. Hashimoto.
\newblock Alpacaeval: An automatic evaluator of instruction-following models, 2023{\natexlab{b}}.
\newblock URL \url{https://github.com/tatsu-lab/alpaca_eval}.

\bibitem[Li et~al.(2023{\natexlab{c}})Li, Bubeck, Eldan, Giorno, Gunasekar, and Lee]{li2023textbooks}
Yuanzhi Li, Sébastien Bubeck, Ronen Eldan, Allie~Del Giorno, Suriya Gunasekar, and Yin~Tat Lee.
\newblock Textbooks are all you need ii: phi-1.5 technical report.
\newblock \emph{arXiv preprint arXiv:2309.05463}, 2023{\natexlab{c}}.
\newblock URL \url{https://arxiv.org/pdf/2309.05463.pdf}.

\bibitem[Mou et~al.(2023)Mou, Ha, Enevoldsen, and Liu]{chenghao_mou_2023_8364980}
Chenghao Mou, Chris Ha, Kenneth Enevoldsen, and Peiyuan Liu.
\newblock Chenghaomou/text-dedup: Reference snapshot, 2023.
\newblock URL \url{https://doi.org/10.5281/zenodo.8364980}.

\bibitem[OpenAI(2023)]{openai2024GPT4}
OpenAI.
\newblock Gpt-4 technical report.
\newblock \emph{arXiv preprint arXiv:2303.08774}, 2023.
\newblock URL \url{https://arxiv.org/pdf/2303.08774.pdf}.

\bibitem[Ponti et~al.(2020)Ponti, Glava, Majewska, Liu, Vuli, and Korhonen]{ponti2020xcopa}
Edoardo~Maria Ponti, Goran Glava, Olga Majewska, Qianchu Liu, Ivan Vuli, and Anna Korhonen.
\newblock Xcopa: A multilingual dataset for causal commonsense reasoning.
\newblock In \emph{Proceedings of the 2020 Conference on Empirical Methods in Natural Language Processing}, 2020.
\newblock URL \url{https://ducdauge.github.io/files/xcopa.pdf}.

\bibitem[Rae et~al.(2022)Rae, Borgeaud, Cai, Millican, Hoffmann, Song, Aslanides, Henderson, Ring, Young, Rutherford, Hennigan, Menick, Cassirer, and et~al]{rae2022scaling}
Jack~W. Rae, Sebastian Borgeaud, Trevor Cai, Katie Millican, Jordan Hoffmann, Francis Song, John Aslanides, Sarah Henderson, Roman Ring, Susannah Young, Eliza Rutherford, Tom Hennigan, Jacob Menick, Albin Cassirer, and Richard~Powell et~al.
\newblock Scaling language models: Methods, analysis and insights from training gopher, 2022.
\newblock URL \url{https://arxiv.org/abs/2112.11446}.

\bibitem[SarvamAI(2023)]{openhathi}
SarvamAI.
\newblock Openhathi series: An approach to build bilingual llms frugally.
\newblock Technical report, SarvamAI, 2023.
\newblock URL \url{https://www.sarvam.ai/blog/announcing-openhathi-series}.

\bibitem[Sutawika et~al.(2023)Sutawika, Gao, Schoelkopf, Biderman, Tow, Abbasi, Fattori, Lovering, Nakhaee, Phang, Thite, Fazz, Aflah, Muennighoff, Wang, Sdtblck, Nopperl, Gakada, Tttyuntian, Researcher2, Chris, Etxaniz, Kasner, Khalid, Hsu, AndyZwei, Ammanamanchi, Groeneveld, Smith, and Tang.]{eval-harness}
Lintang Sutawika, Leo Gao, Hailey Schoelkopf, Stella Biderman, Jonathan Tow, Baber Abbasi, Ben Fattori, Charles Lovering, Farzaneh Nakhaee, Jason Phang, Anish Thite, Fazz, Aflah, Niklas Muennighoff, Thomas Wang, Sdtblck, Nopperl, Gakada, Tttyuntian, Researcher2, Chris, Julen Etxaniz, Zdeněk Kasner, Khalid, Jeffrey Hsu, AndyZwei, Pawan~Sasanka Ammanamanchi, Dirk Groeneveld, Ethan Smith, and Eric Tang.
\newblock A framework for few-shot language model evaluation, 2023.
\newblock URL \url{https://zenodo.org/records/10256836}.

\bibitem[Taori et~al.(2023)Taori, Gulrajani, Zhang, Dubois, Li, Guestrin, Liang, and Hashimoto]{alpaca}
Rohan Taori, Ishaan Gulrajani, Tianyi Zhang, Yann Dubois, Xuechen Li, Carlos Guestrin, Percy Liang, and Tatsunori~B. Hashimoto.
\newblock Stanford alpaca: An instruction-following llama model.
\newblock \url{https://github.com/tatsu-lab/stanford_alpaca}, 2023.

\bibitem[Touvron et~al.(2023)Touvron, Martin, Stone, Albert, Almahairi, Babaei, Bashlykov, Batra, Bhargava, Bhosale, Bikel, Blecher, Ferrer, Chen, Cucurull, Esiobu, Fernandes, Fu, Fu, Fuller, Gao, Goswami, Goyal, Hartshorn, Hosseini, and et~al]{touvron2023llama}
Hugo Touvron, Louis Martin, Kevin Stone, Peter Albert, Amjad Almahairi, Yasmine Babaei, Nikolay Bashlykov, Soumya Batra, Prajjwal Bhargava, Shruti Bhosale, Dan Bikel, Lukas Blecher, Cristian~Canton Ferrer, Moya Chen, Guillem Cucurull, David Esiobu, Jude Fernandes, Jeremy Fu, Wenyin Fu, Brian Fuller, Cynthia Gao, Vedanuj Goswami, Naman Goyal, Anthony Hartshorn, Saghar Hosseini, and et~al.
\newblock Llama 2: Open foundation and fine-tuned chat models.
\newblock \emph{arXiv preprint arXiv:2307.09288}, 2023.
\newblock URL \url{https://arxiv.org/pdf/2307.09288.pdf}.

\bibitem[Vaswani et~al.(2017)Vaswani, Shazeer, Parmar, Uszkoreit, Jones, Gomez, Kaiser, and Polosukhin]{NIPS2017_3f5ee243}
Ashish Vaswani, Noam Shazeer, Niki Parmar, Jakob Uszkoreit, Llion Jones, Aidan~N Gomez, \L~ukasz Kaiser, and Illia Polosukhin.
\newblock Attention is all you need.
\newblock In I.~Guyon, U.~Von Luxburg, S.~Bengio, H.~Wallach, R.~Fergus, S.~Vishwanathan, and R.~Garnett (eds.), \emph{Advances in Neural Information Processing Systems}, 2017.
\newblock URL \url{https://proceedings.neurips.cc/paper_files/paper/2017/file/3f5ee243547dee91fbd053c1c4a845aa-Paper.pdf}.

\bibitem[Wilie et~al.(2020)Wilie, Vincentio, Winata, Cahyawijaya, Li, Lim, Soleman, Mahendra, Fung, Bahar, and Purwarianti]{wilie2020indonlu}
Bryan Wilie, Karissa Vincentio, Genta~Indra Winata, Samuel Cahyawijaya, X.~Li, Zhi~Yuan Lim, S.~Soleman, R.~Mahendra, Pascale Fung, Syafri Bahar, and A.~Purwarianti.
\newblock Indonlu: Benchmark and resources for evaluating indonesian natural language understanding.
\newblock In \emph{Proceedings of the 1st Conference of the Asia-Pacific Chapter of the Association for Computational Linguistics and the 10th International Joint Conference on Natural Language Processing}, 2020.

\bibitem[Winata et~al.(2023)Winata, Aji, Cahyawijaya, Mahendra, Koto, Romadhony, Kurniawan, Moeljadi, Prasojo, Fung, Baldwin, Lau, Sennrich, and Ruder]{winata2022nusax}
Genta~Indra Winata, Alham~Fikri Aji, Samuel Cahyawijaya, Rahmad Mahendra, Fajri Koto, Ade Romadhony, Kemal Kurniawan, David Moeljadi, Radityo~Eko Prasojo, Pascale Fung, Timothy Baldwin, Jey~Han Lau, Rico Sennrich, and Sebastian Ruder.
\newblock Nusax: Multilingual parallel sentiment dataset for 10 indonesian local languages, 2023.
\newblock URL \url{https://arxiv.org/abs/2205.15960}.

\bibitem[Wu et~al.(2016)Wu, Schuster, Chen, Le, Norouzi, Macherey, Krikun, Cao, Gao, Macherey, Klingner, Shah, Johnson, Liu, Łukasz Kaiser, Gouws, Kato, Kudo, Kazawa, Stevens, Kurian, Patil, Wang, Young, Smith, Riesa, Rudnick, Vinyals, Corrado, Hughes, and Dean]{google-translate}
Yonghui Wu, Mike Schuster, Zhifeng Chen, Quoc~V. Le, Mohammad Norouzi, Wolfgang Macherey, Maxim Krikun, Yuan Cao, Qin Gao, Klaus Macherey, Jeff Klingner, Apurva Shah, Melvin Johnson, Xiaobing Liu, Łukasz Kaiser, Stephan Gouws, Yoshikiyo Kato, Taku Kudo, Hideto Kazawa, Keith Stevens, George Kurian, Nishant Patil, Wei Wang, Cliff Young, Jason Smith, Jason Riesa, Alex Rudnick, Oriol Vinyals, Greg Corrado, Macduff Hughes, and Jeffrey Dean.
\newblock Google's neural machine translation system: Bridging the gap between human and machine translation, 2016.
\newblock URL \url{https://arxiv.org/abs/1609.08144}.

\bibitem[Zhang et~al.(2022)Zhang, Hashimoto, Wan, Liu, Liu, Xiong, and Yu]{zhang2022pretrained}
Jianguo Zhang, Kazuma Hashimoto, Yao Wan, Zhiwei Liu, Ye~Liu, Caiming Xiong, and Philip~S. Yu.
\newblock Are pretrained transformers robust in intent classification? a missing ingredient in evaluation of out-of-scope intent detection, 2022.
\newblock URL \url{https://arxiv.org/abs/2106.04564}.

\bibitem[Zheng et~al.(2023)Zheng, Chiang, Sheng, Li, Zhuang, Wu, Zhuang, Li, Lin, Xing, Gonzalez, Stoica, and Zhang]{zheng2023lmsyschat1m}
Lianmin Zheng, Wei-Lin Chiang, Ying Sheng, Tianle Li, Siyuan Zhuang, Zhanghao Wu, Yonghao Zhuang, Zhuohan Li, Zi~Lin, Eric.~P Xing, Joseph~E. Gonzalez, Ion Stoica, and Hao Zhang.
\newblock Lmsys-chat-1m: A large-scale real-world llm conversation dataset, 2023.
\newblock URL \url{https://arxiv.org/abs/2309.11998}.

\bibitem[Üstün et~al.(2024)Üstün, Aryabumi, Yong, Ko, D'souza, Onilude, Bhandari, Singh, Ooi, Kayid, Vargus, Blunsom, Longpre, Muennighoff, Fadaee, Kreutzer, and Hooker]{aya-model}
Ahmet Üstün, Viraat Aryabumi, Zheng-Xin Yong, Wei-Yin Ko, Daniel D'souza, Gbemileke Onilude, Neel Bhandari, Shivalika Singh, Hui-Lee Ooi, Amr Kayid, Freddie Vargus, Phil Blunsom, Shayne Longpre, Niklas Muennighoff, Marzieh Fadaee, Julia Kreutzer, and Sara Hooker.
\newblock Aya model: An instruction finetuned open-access multilingual language model.
\newblock \emph{arXiv preprint arXiv:2402.07827}, 2024.
\newblock URL \url{https://arxiv.org/pdf/2402.07827.pdf}.

\end{thebibliography}

\clearpage
\newpage
\appendix

\section{Appendix}
\subsection{Functions used for Evaluation of Generated Output by Models}

\subsubsection{Example of Evaluation function used for IndoMMLU}

\begin{lstlisting}

def evaluate_correctness(row):
    output = str(row['Output']).lower()
    answer = str(row['answer']).lower()
    Input = str(row['Input'])
    
    if len(output) == 1:
        return output[0] == answer
    elif len(output) > 1 and (output[1] == '.' ):
        return output[0] == answer
    else:
        return GPT_4_evaluator(Input, output, answer)
        
\end{lstlisting}

\subsubsection{Example of Evaluation function used for ID-EN}

\begin{lstlisting}

def evaluate_correctness(row):
    answer = str(row['answer']).strip()
    output = str(row['Output_Mapped']).strip()
    if output in ['1','0']:
        return output[0]==answer
    else:
        return GPT_4_evaluator(output,answer)
        
\end{lstlisting}

\subsubsection{Example of Evaluation function used for X-Copa-ID}

\begin{lstlisting}
def evaluate_correctness(row):
    answer = str(row['answer'])
    output = str(row['Output'])
    if answer.lower() in output.lower():
        return True
    elif 'Saya tidak dapat menemukan jawaban atas pertanyaan 
    yang diajukan.'.lower() in output.lower():
        return False
    else:
        return GPT_4_evaluator(output,answer)
\end{lstlisting}

\subsubsection{Example of Evaluation function used for Intent-Classification}

\begin{lstlisting}
def check_occurrence(sentence, words_set):
    count = sum(1 for word in words_set if word.lower() in sentence.lower())
    return count >= 2

def return_final_output_intent(output, negative_intent="tidak ada"):
    if isinstance(output, float):
        output = str(output)
    intent_list = [
        'automatic top up', 'balance not updated after cheque or 
        cash deposit',
        'declined card payment', 'declined transfer',
        'edit personal details',
    ]

    if check_occurrence(output, intent_list):
        return negative_intent

    for expected_intent in intent_list:
        if expected_intent.lower() in output.lower():
            return expected_intent.lower()

    return negative_intent
\end{lstlisting}

\subsubsection{Example of Evaluation function used for Colloquial Detection}

\begin{lstlisting}
def check_occurrence(sentence, words_set):
    count = sum(1 for word in words_set if word.lower() in sentence.lower())
    return count >= 2
    
def return_in_format(response):
    if response is None or isinstance(response, (int, float)):
        return -1
    
    words_set = ("ceremonial", "polished", "everyday", "conversational", 
    "colloquial")

    if check_occurrence(response, words_set):
        return -1
    elif any(word in response.lower() for word in ("ceremonial", "polished", 
    "everyday")):
        return 0
    elif any(word in response.lower() for word in ("conversational", 
    "colloquial")):
        return 1
    else:
        return response.lower()
\end{lstlisting}

\subsubsection{Example of Evaluation function used for NusaX-Senti}

\begin{lstlisting}
dictionary = {
    'positive': 'positif',
    'negative': 'negatif',
    'neutral': 'netral'
}

def evaluate(output, answer):
    output = output.replace('.', '')
    if ' ' not in output:
        output_lower = output.lower()
        answer_lower = answer.lower()
        
        if output_lower == answer_lower:
            return True
        elif output_lower in dictionary:
            return dictionary[output_lower].lower() == answer_lower
        else:
            return False
    else:
        result = GPT_4_evaluator(output, answer)
        return result
        
\end{lstlisting}

\subsubsection{Example of Evaluation function used for ID-HateSpeech}

\begin{lstlisting}
def evaluate_correctness(row):
    answer = str(row['answer']).strip()
    output = str(row['Output']).strip()
    output = output.replace('.','')
    if len(output) == 1:
        return output == answer
    elif output[0] == '1' or output[0] == '0':
        return output[0] == answer
    else:
        return GPT_4_evaluator(output, answer)
\end{lstlisting}

\subsubsection{Example of Evaluation function used for TydiQA-ID}

\begin{lstlisting}
def evaluate_correctness(row):
    answer = str(row['answer'])
    output = str(row['Output'])
    if answer.lower() in output.lower():
        return True
    elif 'Saya tidak dapat menemukan jawaban atas pertanyaan 
    yang diajukan.'.lower() in output.lower():
        return False
    else:
        return GPT_4_evaluator(output,answer)
\end{lstlisting}

\subsubsection{Example of Evaluation function used for IndoSum}

\begin{lstlisting}
from rouge import Rouge
rouge = Rouge()
def calculate_rouge_l(row):
    scores = rouge.get_scores(str(row['Output']), 
    str(row['answer']))[0]['rouge-l']
    return scores['r'], scores['p'], scores['f']
\end{lstlisting}

\subsection{Prompts Used for Evaluation in GPT-4-0613}

\subsubsection{Evaluation Prompt for IndoMMLU}

{\itshape
Given the following options:\{Options\}.The model's generated response is:\{output_text\}.The correct answer is: \{answer\}.Your task is to check if the model's response is correct or not? Provide a response with Yes or No only.
}

\subsubsection{Evaluation Prompt for X-Copa-ID}

{\itshape Your task is to check if the Actual Answer is present in the Generated Answer.Generated Answer:\{generated_answer\},Actual Answer:\{actual_answer\}.Provide a response with Yes or No only.}

\subsubsection{Evaluation Prompt for TydiQA-ID}

{\itshape Your task is to check if the Actual Answer is present in the Generated Answer.Generated Answer:\{generated_answer\},Actual Answer:\{actual_answer\}.Provide a response with Yes or No only.}

\subsubsection{Evaluation Prompt for ID-HateSpeech}

{\itshape Your task is to Verify if the given output is same as expected answer. Output: \{output_text\}, Expected Answer: \{expected_answer\}.Provide a response with Yes or No only.}

\subsubsection{Evaluation Prompt for NusaX-Senti}

{\itshape Your task is to Verify if the given output is same as expected answer. Output: \{output_text\}, Expected Answer: \{expected_answer\}.Provide a response with Yes or No only.}

\subsection{Qualitative Samples for Downstream Tasks}

\subsubsection{Caption Generation}

\textbf{Input:}
{\itshape \\
"Di bawah ini adalah instruksi yang menjelaskan tugas. Tulis respons yang secara tepat melengkapi permintaan. \\
\#\#\#Instruksi: \\
Buatkan caption untuk post Instagram dengan tema ""bunga putih"" \\
\#\#\#Respons:" 
}

\textbf{Input(Translated):}
{\itshape \\
"Below are instructions that explain the task. Write a response that precisely completes the request. \\
\#\#\# Instruction: \\
Make a caption for Instagram post with the theme ""White Flower"" \\
\#\#\# Response:" \\
}

\textbf{Llama-2-7B-Finetuned Output:}
{\itshape \\
Sensasi keindahan bunga putih yang memikat hati dan menenangkan pikiran. Nikmati keindahan alam dalam setiap detik.
}

\textbf{Aya-101:}
{\itshape \\
Sebuah buket bunga putih yang indah, sempurna untuk hari yang cerah!
}

\textbf{Mixtral-8x7B-Instruct:}
{\itshape \\
 Delicate beauty in monochrome, a serene touch of pure white flowers. \#bungaputih \#floralperfection
 }

\textbf{Komodo-7B-Instruct:}
{\itshape \\
Keindahan bunga putih yang mempesona, mengingatkanku pada kesucian dan ketulusan hati. \#bungaputih \#keindahanalam \#kesucian
}

\subsubsection{Email Generation}

\textbf{Input:}
{\itshape \\
"Di bawah ini adalah instruksi yang menjelaskan tugas. Tulis respons yang secara tepat melengkapi permintaan. \\
\#\#\# Instruksi: \\
Tuliskan balasan email yang sopan untuk menolak tawaran pekerjaan yang diberikan karena renumerasi yang ditawarkan terlalu kecil, namun dengan bahasa yang halus. \\
\#\#\# Respons:"
}

\textbf{Input(Translated):}
{\itshape \\
"Below are instructions that explain the task. Write a response that precisely completes the request. \\
\#\#\# Instruction: \\
Write a polite email reply to refuse the job offer given because the renumeration offered is too small, but with a subtle language. \\
\#\#\# Response:"
}

\textbf{Llama-2-7B-Finetuned Output:}
{\itshape \\
Kepada [Nama Pemberi Tawaran Pekerjaan], \\\\
Saya sangat menghargai kesempatan yang diberikan untuk melamar pekerjaan di perusahaan Anda. Namun, saya ingin menyampaikan bahwa saya harus menolak tawaran pekerjaan ini karena renumerasi yang ditawarkan terlalu kecil untuk memenuhi kebutuhan hidup saya dan keluarga.\\\\
Saya sangat mengapresiasi kesabaran dan kerja keras tim Anda dalam mencari kandidat yang tepat untuk posisi ini. Namun, saya harus mempertimbangkan faktor keuangan dan kesejahteraan saya sendiri dan keluarga saya. Oleh karena itu, saya harus menolak tawaran pekerjaan ini dan mengucapkan terima kasih atas waktu dan usaha yang telah diberikan oleh tim Anda.\\\\
Saya berharap dapat menemukan kesempatan pekerjaan yang lebih sesuai dengan kemampuan dan kebutuhan saya di masa depan. Terima kasih atas perhatian Anda.\\\\
Salam hormat \\\\
Nama Anda
}

\textbf{Aya-101:}
{\itshape \\
Halo, terima kasih atas tawaran Anda. Saya senang mendengar bahwa Anda tertarik pada pekerjaan ini. Namun, saya pikir renumerasi yang Anda tawarkan terlalu kecil untuk saya. Saya pikir saya lebih cocok untuk pekerjaan yang lebih besar dengan gaji yang lebih tinggi. Jika Anda memiliki kesempatan untuk mempertimbangkan saya, saya akan senang untuk menghubungi Anda kembali. Terima kasih atas waktu dan perhatiannya. Sejujurnya, [Nama Anda]
}

\textbf{Mixtral-8x7B-Instruct:}
{\itshape \\
Subject: Regarding Your Job Offer \\\\
Dear [Hiring Manager's Name], \\\\
I hope this email finds you well. I am writing in response to the job offer that I received from your esteemed organization. I am truly honored to have been considered for this position and appreciate the time and effort invested in the recruitment process. \\\\
However, after careful consideration, I regret to inform you that I am unable to accept the position at this time. Although I am excited about the opportunity to contribute to [Company Name] and work alongside a talented team of professionals, I have come to the conclusion that the proposed compensation package does not align with my financial needs and career goals. \\\\
I want to emphasize that this decision was not made lightly, and I am grateful for the consideration. I am genuinely impressed with [Company Name]'s mission, vision, and culture, and I am confident that the organization will continue to succeed and thrive in the industry.\\\\
I would like to express my appreciation to you and the entire team for your time and support throughout the recruitment process. I hope that we can keep the lines of communication open for future opportunities that may be a better fit for both parties.\\
Thank you again for the opportunity, and I wish you and the entire team all the best.\\
Sincerely, \\\\
Your Name
 }

\textbf{Komodo-7B-Instruct:}
{\itshape \\
Kepada Tim Rekrutmen, \\\\
Terima kasih atas kesempatan yang diberikan untuk mempertimbangkan tawaran pekerjaan yang diberikan. Saya sangat mengapresiasi kesempatan ini dan sangat tertarik dengan posisi yang ditawarkan. \\\\
Namun, setelah mempertimbangkan dengan seksama tawaran ini, saya ingin memberitahukan bahwa saya dengan sangat menyesal harus menolak tawaran ini. Sayangnya, renumerasi yang ditawarkan terlalu kecil untuk memenuhi ekspektasi saya dalam hal kompensasi dan manfaat yang diberikan.\\\\
Saya ingin mengucapkan terima kasih atas waktu dan usaha yang telah diberikan oleh tim rekrutmen dalam proses rekrutmen ini. Saya berharap kesuksesan dan keberhasilan untuk perusahaan di masa depan. \\\\
Salam hormat, \\\\
Your Name
}

\subsubsection{Suggestion Generation}

\textbf{Input:} 
{\itshape\\
"Di bawah ini adalah instruksi yang menjelaskan tugas. Tulis respons yang secara tepat melengkapi permintaan.\\
\#\#\# Instruksi: \\
Gue mau nembak gebetan gue di pantai, kasih tips dong supaya bisa romantis pisan \\
\#\#\# Respons:" 
}

\textbf{Input(Translated):}
{\itshape \\
"Below are instructions that explain the task. Write a response that precisely completes the request. \\
\#\#\# Instruction: \\
I want to confess to my crush at a beach, give me some tips to make it romantic \\
\#\#\# Response:" \\
}

\textbf{Llama-2-7B-Finetuned Output:}
{\itshape \\
Anjeun kudu nyiapkeun tempat anu alus pikeun némbak, biasana tempat anu cerah sareng sepi. Anjeun oge kudu mastikeun yén anjeun boga bahan pikeun némbak, saperti bal, papan némbak, jeung papan pikeun némbak. Sanggeus anjeun manggihan tempat anu alus pikeun némbak, anjeun kudu ngabahas jeung babaturan anjeun ngeunaan naon anu anjeun hoyong ngalakukeun. Contona, anjeun bisa ngomong, "Kuring bade némbak bal kuring kahiji". Lamun babaturan anjeun teu boga ideu, anjeun bisa ngomong, "Kuring bade némbak bal kuring kahiji". Sanggeus anjeun geus nyieun rencana, anjeun kudu ngabahas jeung babaturan anjeun ngeunaan cara ngalakukeunana. Contona, anjeun bisa ngomong,....
}

\textbf{Aya-101:}
{\itshape \\
Pastikan cuaca nyaman dan bebas gangguan.
}

\textbf{Mixtral-8x7B-Instruct:}
{\itshape \\
Selamat malam! Sungguh menyenangkan dengan tujuan liburan romancemu ke pantai. Berikut adalah beberapa tips untuk membuat gebetanmu menjadi romantis: \\
1. Pilih waktu: Pilih waktu sore hari atau malam hari ketika cuaca sedang sejuk dan angin menyapu pantai. Warnanya akan menjadi lebih terang, dan keindahan pasir putih akan menjadi lebih menonjol. \\
2. Temukan tempat yang romantis: Cari sebuah tempat yang tenang dan tenang, seperti area yang jauh dari keramaian. Temukan area pasir yang lembut dan bersih, dan jika mungkin, dekat dengan air laut. \\
3. Bring the essentials: Jangan lupa untuk membawa bahan-bahan untuk membuat gebetan yang romantis, seperti lampu tengah malam, selimut lembut, dan bantal. Kamu juga mungkin ingin membawa beberapa makanan ringan dan minuman untuk menikmati bersama. \\
4. Dress up: Pilih pakaian yang nyaman dan romantis. Pilih warna yang cerah dan berkilau, seperti biru muda atau hijau muda, yang akan membuatmu terlihat cantik di bawah sinar matahari terbenam. \\
5. Take your time: Jangan terlalu cepat membuat gebetan. Nikmati saat itu, bersantai, dan main-main di pasir. Ingat, liburan romance ini adalah tentang kalian dua, bukan hanya tentang gebetan. \\
6. Berbagi cinta: Sebelum memulai gebetan, berbagi cinta dengan pasanganmu. B"
 }

\textbf{Komodo-7B-Instruct:}
{\itshape \\
1. Pilih waktu yang tepat. Pilihlah waktu yang tepat untuk mengajak gebetanmu ke pantai. Pilihlah waktu yang tenang dan sunyi, sehingga kamu bisa berdua lebih nyaman dan santai. \\
2. Siapkan makanan ringan. Siapkan makanan ringan seperti buah-buahan segar, kue-kue kecil, atau camilan lainnya yang bisa dinikmati saat berada di pantai. Ini akan membuat suasana semakin romantis. \\
3. Siapkan lampu lilin. Siapkan lampu lilin yang bisa dipasang di sekitar pantai. Ini akan memberikan suasana yang lebih romantis dan intim.  \\
4. Siapkan musik. Siapkan musik yang romantis dan sesuai dengan suasana pantai. Ini akan membuat suasana semakin intim dan romantis. \\
5. Siapkan hadiah kecil. Siapkan hadiah kecil untuk diberikan kepada gebetanmu. Ini akan membuat suasana semakin romantis dan intim. \\
6. Jangan lupakan tatapan mata. Jangan lupa untuk tatap mata dan berinteraksi dengan gebetanmu. Ini akan membuat suasana semakin romantis dan intim. \\
7. Nikmati momen bersama. Nikmati momen bersama dengan gebetanmu di pantai. Ini akan membuat suasana semakin romantis dan intim. \\
}

\subsubsection{Suggestion Generation}

\textbf{Input:}
{\itshape \\
"Di bawah ini adalah instruksi yang menjelaskan tugas. Tulis respons yang secara tepat melengkapi permintaan.\\
\#\#\# Instruksi: \\
Tolong berikan beberapa ide untuk games yang bisa dilakukan keluarga saat chinese new year \\
\#\#\# Respons:"
}

\textbf{Input(Translated):}
{\itshape \\
"Below are instructions that explain the task. Write a response that precisely completes the request. \\
\#\#\# Instruction: \\
Please give some ideas for games that family can do during Chinese New Year \\
\#\#\# Response:"
}

\textbf{Llama-2-7B-Finetuned Output:}
{\itshape \\
1. Game kue tradisional: Keluarga dapat membuat kue tradisional seperti kue tahu, kue pisang, atau kue ubi jalar dan bermain bersama-sama. \\
2. Game kartu: Keluarga dapat bermain game kartu tradisional seperti mahjong atau weiqi.\\
3. Game tebak-tebakan: Keluarga dapat bermain game tebak-tebakan dengan tema Cina seperti game Mahjong atau Go. \\
4. Game karaoke: Keluarga dapat menyanyikan lagu-lagu Cina tradisional dan menonton orang lain menyanyikan lagu-lagu tersebut. \\
5. Game memasak: Keluarga dapat memasak makanan tradisional Cina bersama-sama dan bermain game memasak. \\
6. Game tebak-tebakan: Keluarga dapat bermain game tebak-tebakan dengan tema Cina seperti game Mahjong atau Go. \\
7. Game papan: Keluarga dapat bermain game papan tradisional seperti Go atau Shogi dengan tema Cina. \\
8. Game permainan papan: Keluarga dapat bermain game permainan papan tradisional seperti Go atau Shogi dengan tema Cina. \\
9. Game tebak-tebakan: Keluarga dapat bermain game tebak-tebakan dengan tema Cina seperti game Mahjong atau Go.\\
10. Game karaoke: Keluarga dapat menyanyikan lagu-lagu Cina tradisional dan menonton orang lain menyanyikan lagu-lagu tersebut.
}

\textbf{Aya-101:}
{\itshape \\
Berikut adalah beberapa ide untuk permainan yang bisa dilakukan keluarga saat chinese new year: - Keluarga bisa bermain catur. - Keluarga bisa bermain catur. - Keluarga bisa bermain catur. - Keluarga bisa bermain catur. - Keluarga bisa bermain catur. - Keluarga bisa bermain catur. - Keluarga bisa bermain catur. - Keluarga bisa bermain catur. - Keluarga bisa bermain catur. - Keluarga bisa bermain catur. - Keluarga bisa bermain catur. - Keluarga bisa bermain .... 
}

\textbf{Mixtral-8x7B-Instruct:}
{\itshape \\
1. ""Lucky Red Envelope Hunt"": This game involves hiding red envelopes (ang pao) around the house, and having family members search for them. The envelopes can contain chocolate coins or small toys, adding to the excitement. \\
2. ""Chinese New Year Trivia"": Create a trivia game with questions about Chinese New Year traditions, history, and culture. This can be a fun and educational way to engage the whole family. \\
3. ""Chinese Character Practice"": Set up a station where family members can practice writing Chinese characters using traditional calligraphy brushes and ink. This can be a relaxing and meditative activity. \\
4. ""Lantern Making"": Provide materials for family members to create their own lanterns, which can be hung up around the house as decorations. This can be a fun and creative activity for all ages. \\
5. ""Chinese Zodiac Scavenger Hunt"": Create a scavenger hunt where family members have to find objects or pictures that represent each of the 12 animals in the Chinese zodiac.\\
6. ""Chinese Cooking Challenge"": Have a cooking competition where family members have to create a dish using traditional Chinese New Year ingredients, such as dumplings, fish, or noodles. \\
7. ""Chinese Tea Ceremony"": Hold a traditional Chinese tea ceremony, where family members can learn about the history and significance of tea in Chinese culture. \\
8. ""Chinese Knot Tying"": Teach family members how to tie traditional Chinese knots using ribbons or cords. This can be a fun and challenging activity that results in beautiful decorations. \\
9. ""Chinese Chess or Mahjong Tournament"": Organize a tournament where family members can play traditional Chinese board games like chess or Mahjong. \\
10. ""Lion Dance Performance"": Encourage family members to create their own lion dance costumes and put on a performance. This can be a fun and energetic way to celebrate Chinese New Year traditions. \\
 }

\textbf{Komodo-7B-Instruct:}
{\itshape \\
1. Permainan tebak-tebakan \\
2. Permainan kartu seperti Uno, Blackjack, atau Poker \\
3. Permainan papan seperti Monopoli atau Risk \\
4. Permainan video seperti Mario Kart atau Mario Party \\
5. Membuat kue bersama-sama \\
6. Membuat hiasan atau dekorasi rumah bersama-sama \\
7. Mengadakan pesta kostum dan foto bersama \\
8. Mengadakan permainan bola atau olahraga bersama \\
9. Mengadakan acara memasak bersama-sama \\
10. Mengadakan acara menonton film bersama-sama. \\
}

\subsubsection{Translation}

\textbf{Input:}
{\itshape \\
"Di bawah ini adalah instruksi yang menjelaskan tugas. Tulis respons yang secara tepat melengkapi permintaan. \\
\#\#\# Instruksi: \\
apa arti kata ""dahar"" dalam bahasa sunda? \\
\#\#\# Respons:" \\
}

\textbf{Input(Translated):}
{\itshape
"Below are instructions that explain the task. Write a response that precisely completes the request. \\
\#\#\# Instruction: \\
What does the word ""dahar"" mean in Sundanese? \\
\#\#\# Response:"
}

\textbf{Llama-2-7B-Finetuned Output:}
{\itshape \\
Dahar nyaéta kecap anu dipaké pikeun ngagambarkeun kagiatan ngadahar.
}

\textbf{Aya-101:}
{\itshape \\
Dahar hartina dahar
}

\textbf{Mixtral-8x7B-Instruct:}
{\itshape   \\ 
 Hallo! Kata "dahar" dalam Bahasa Sunda memiliki arti 'perut atau bagian bawah badan'. Semoga dapat membantu Anda!
 }

\textbf{Komodo-7B:}
{\itshape \\
Dahar artinya makan.
}

\end{document}